\documentclass[]{fairmeta}

\title{\methodname{}: Video Auto Reasoning via Thinking Once, Answering Twice}

\author[1,2,*]{Shuming Liu}
\author[1,2]{Mingchen Zhuge}
\author[1]{Changsheng Zhao}
\author[1]{Jun Chen}
\author[1]{Lemeng Wu}
\author[1]{\\Zechun Liu}
\author[1]{Chenchen Zhu}
\author[1]{Zhipeng Cai}
\author[1]{Chong Zhou}
\author[1,2]{Haozhe Liu}
\author[1]{Ernie Chang}
\author[1]{Saksham Suri}
\author[1]{Hongyu Xu}
\author[1]{Qi Qian}
\author[1]{Wei Wen}
\author[1]{Balakrishnan Varadarajan}
\author[3]{Zhuang Liu}
\author[1]{Hu Xu}
\author[1]{Florian Bordes}
\author[1]{\\Raghuraman Krishnamoorthi}
\author[2,\dagger]{Bernard Ghanem}
\author[1,\dagger]{Vikas Chandra}
\author[1,\dagger]{Yunyang Xiong}

\affiliation[1]{Meta AI}
\affiliation[2]{King Abdullah University of Science and Technology (KAUST)}
\affiliation[3]{Princeton University}

\contribution[*]{Work done at Meta}
\contribution[\dagger]{Project lead}

\usepackage{pifont} 
\newcommand{\cmark}{\ding{51}}%
\newcommand{\xmark}{\ding{55}}%
\newcommand{\methodname}{VideoAuto-R1}

\usepackage{makecell}
\usepackage{amsfonts}
\usepackage{amsmath}
\usepackage{amssymb}
\usepackage[dvipsnames]{xcolor}
\usepackage[dvipsnames]{colortbl}
\usepackage{arydshln}
\usepackage{algorithm} 
\usepackage[most]{tcolorbox}
\usepackage{graphicx}
\usepackage{booktabs}
\usepackage{multirow}
\usepackage{caption}
\usepackage{threeparttable}
\usepackage{algpseudocode}
\usepackage{listings}
\usepackage{bm}
\usepackage{hhline}
\usepackage{arydshln}
\usepackage{array,etoolbox}
\usepackage{wrapfig}

\def\eg{\textit{e.g.}}
\def\ie{\textit{i.e.}}

\abstract{
Chain-of-thought (CoT) reasoning has emerged as a powerful tool for multimodal large language models on video understanding tasks. However, its necessity and advantages over direct answering remain underexplored. In this paper, we first demonstrate that for RL-trained video models, direct answering often matches or even surpasses CoT performance, despite CoT producing step-by-step analyses at a higher computational cost. Motivated by this, we propose \textbf{\methodname{}}, a video understanding framework that adopts a \textit{``reason-when-necessary''} strategy. During training, our approach follows a \textbf{Thinking Once, Answering Twice} paradigm: the model first generates an initial answer, then performs reasoning, and finally outputs a reviewed answer. Both answers are supervised via verifiable rewards. During inference, the model uses the confidence score of the initial answer to determine whether to proceed with reasoning. Across video QA and grounding benchmarks, \methodname{} achieves state-of-the-art accuracy with significantly improved efficiency, reducing the average response length by $\sim$3.3x, \eg, from 149 to just 44 tokens. Moreover, we observe a low rate of thinking-mode activation on perception-oriented tasks, but a higher rate on reasoning-intensive tasks. This suggests that explicit language-based reasoning is generally beneficial but not always necessary.
}

\correspondence{\email{shuming.liu@kaust.edu.sa}, \email{yunyang@meta.com}}

\metadata[Project \& Demo]{\url{https://ivul-kaust.github.io/projects/videoauto-r1}}

\begin{document}

\maketitle

\section{Introduction}
\label{sec:intro}

Recent advances in explicit reasoning, most notably chain-of-thought (CoT)~\citep{sahoo2024systematic}, have pushed large language models (LLMs) and multimodal LLMs to new heights~\citep{team2023gemini,jaech2024openai,guo2025deepseek,xu2025toward}. These models often operate in a \textit{thinking-mode}, which generates an explicit, step-by-step CoT to analyze the problem, verify intermediate conclusions, and revise them as necessary. On text-only tasks such as mathematics and coding, reasoning models markedly improve problem-solving capabilities~\citep{shao2024deepseekmath,guo2025deepseek}. In the image domain, many works also aim to enhance both perceptual understanding and complex visual reasoning~\citep{yang2025r1,wang2025vl,wang2025sota,zheng2025deepeyes}. Recently, video reasoning has also drawn substantial attention~\citep{chen2025scaling,li2025veripo,li2025videochat,fu2025love}. These methods encourage extended thinking traces that analyze frames and events in detail~\citep{wang2025time,ghazanfari2025chain}, retrieve relevant spatial objects~\citep{gong2025reinforcing}, reason about temporal order~\citep{feng2025video,dang2025reinforcing}, and call external tools~\citep{zhang2025thinking,xie2025video}, substantially improving models’ performance on video QA and temporal grounding tasks.

However, unlike math problems where inputs are symbolic and noise-free, video understanding naturally focuses more on visual perception than on explicit step-by-step thinking. Once the perception is accurate or confirmed, the remaining symbolic reasoning tends to be shallow. This raises an important question: \textit{Is complex reasoning always necessary for general video understanding?} To investigate, we analyze existing models and uncover a surprising pattern: for RL-trained video reasoning models, a direct-answer strategy, \ie, providing a final answer without explanations, often matches, and sometimes even outperforms, thinking-mode inference (see Table~\ref{table:think_direct_analysis}). Only on benchmarks that explicitly demand multi-step reasoning, \eg, VideoMMMU~\citep{hu2025video}, CoT shows a consistent advantage. This finding suggests that long reasoning traces for video tasks \textbf{do not} inherently improve accuracy and may even cause overthinking that degrades performance. Similar phenomena have also been observed in the text and image domains~\citep{sui2025stop,kumar2025overthink}. 

Another issue of the always-thinking strategy is lower efficiency~\citep{chen2024not,qu2025survey,li2025adaptive}. Thinking-only models typically generate long responses with hundreds of tokens, while direct answering often requires much fewer tokens. Given the autoregressive nature of LLMs, these longer traces substantially increase latency and inference cost. Therefore, an efficient and effective approach to video reasoning is to reason only when necessary, that is, to employ \textbf{\textit{auto-thinking}}.

Auto-thinking, or adaptive reasoning, allows a model to decide whether to answer directly or to invoke CoT reasoning based on input complexity~\citep{yang2025qwen3,cheng2025incentivizing,lou2025adacot}. Prior work has focused on text and images, typically learning a switching policy via supervised fine-tuning (SFT) or reinforcement learning (RL) to dynamically select the thinking mode~\citep{zhang2025adaptthink,yang2025r,xie2025arm2}. 
Extending these strategies directly to video is non-trivial: the correlation between explicit reasoning and accuracy is weak in video due to visual ambiguity and long-range temporal noise. Moreover, truly \emph{must-think} video samples are relatively rare, which necessitates careful data curation during training~\citep {zhan2025kat}. In our early experiments (Table~\ref{table:autothink_ablation}), rigidly enforcing think/no-think decisions \textit{during training} often led to model collapse (always think or no-think) and poor generalization at test time.

To enable video auto-thinking that reasons only when necessary, we propose a thinking once, answering twice mechanism. Instead of optimizing a binary objective (think or no-think) for each sample, we introduce a new response template: \textcolor{blue}{\textit{answer → think → answer}} (see Table~\ref{table:visionthink_prompt} for the full prompt). During training, the model first provides an initial answer, then performs explicit reasoning, and finally outputs a reviewed answer. Both answers are supervised with verifiable rewards, with a larger weight assigned to the final answer to encourage the model to refine or confirm its initial answer. Notably, this paradigm eliminates the need for manual think/no-think labeling during training; the model simply learns to make both answers correct. As a result, the response can always begin with a short, direct answer, followed by a step-by-step explanation.

\begin{figure}[t]
\centering
\includegraphics[width=0.985\linewidth]{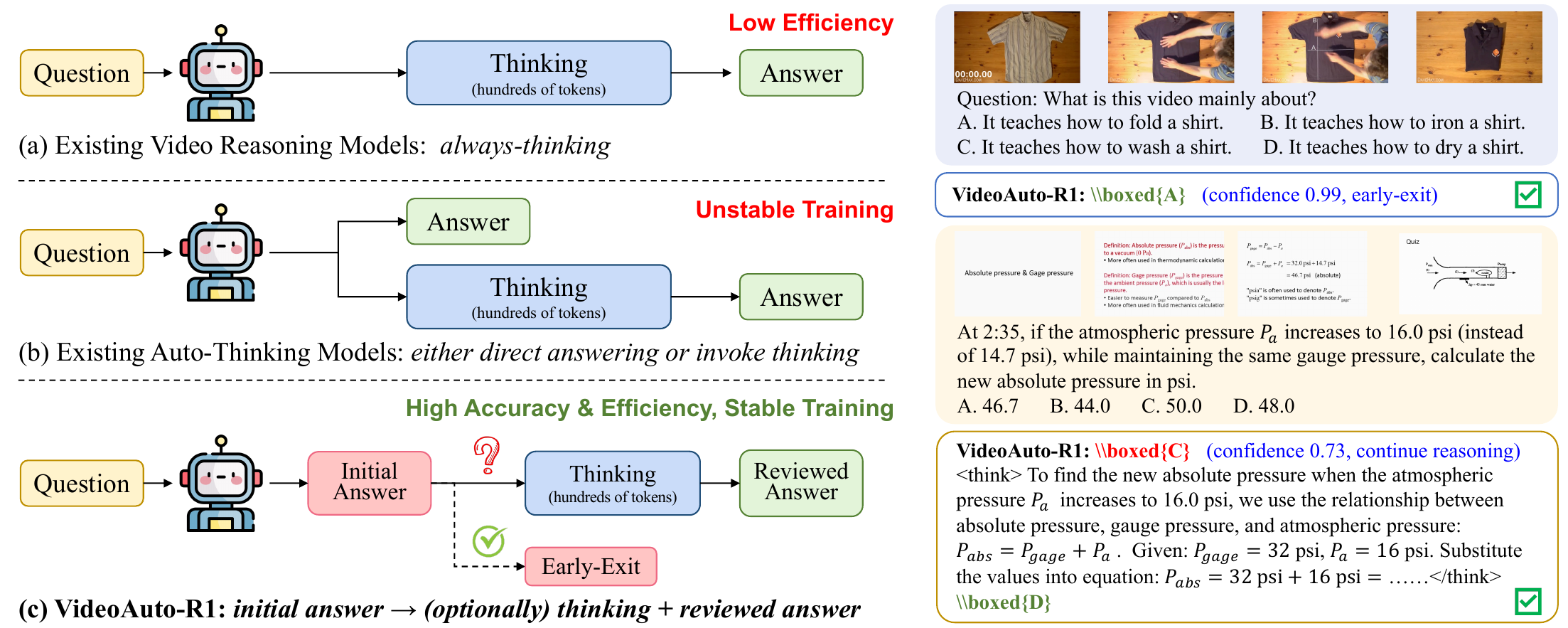}
\vspace{-3pt}
\caption{\textbf{\methodname{} follows a thinking once, answering twice paradigm}. In training, both the initial answer and the reviewed answer are supervised with verifiable rewards. During inference, an early-exit mechanism is adopted to dynamically determine whether to proceed with CoT reasoning. Robot icon from \cite{robot_icon}.}
\label{fig:intro}
\vspace{-5pt}
\end{figure}

At inference time, rather than relying on an additional mode switch token or head, we employ a simple rule-based early-exit strategy. After the model outputs the first answer, we compute the length-normalized mean log probability of those answer tokens as the confidence score. If it exceeds a threshold, we treat the initial answer as sufficiently reliable and terminate the decoding early, equivalent to direct answering~\citep{yue2025don}. Otherwise, the model continues to generate the reasoning trace and the reviewed answer. Thus, the thinking-mode activation is solely determined \textit{at test time}. Empirically, as shown in Table~\ref{table:dataset_prob}, the confidence score correlates well with mode-switch accuracy, allowing us to precisely determine which samples require reasoning. We refer to the resulting training and inference framework as \textbf{\methodname{}} (see Figure~\ref{fig:intro}).

Evaluations across benchmarks reveal two key advantages of \methodname{}. \textbf{(1) Accuracy:} for challenging inputs that benefit from step-by-step reasoning, the model reliably activates thinking mode, refines its initial answer, and achieves state-of-the-art performance; \textbf{(2) Efficiency:} for inputs that do not require reasoning, early-exit suppresses unnecessary token generation, reducing latency and inference cost compared to standard video reasoning models. Notably, on perception-oriented benchmarks such as MVBench~\citep{li2024mvbench}, the think-mode activation rate is low (25\%), while on reasoning-intensive benchmarks such as VideoMMMU~\citep{hu2025video}, it rises to 51\%. Overall, \methodname{} reduces the average response length from 149 to just 44 tokens while preserving accuracy. We summarize our contributions as follows:

\begin{enumerate}
\item To the best of our knowledge, we present the first systematic study showing that existing video reasoning models perform comparably in direct and CoT modes, cautioning against unconditional reliance on CoT given its high computation cost and modest gains.

\item We propose \methodname{}, which couples a thinking once, answering twice training paradigm with a confidence-based early-exit inference strategy. It eliminates the need for per-sample think/no-think labels, yielding a simple yet effective adaptive reasoning model.

\item Through extensive experiments and ablations, we show that \methodname{} achieves state-of-the-art accuracy while substantially improving efficiency across video QA and temporal grounding tasks.

\end{enumerate}

\section{Related Work}

\subsection{Chain-of-Thought Reasoning}
Chain-of-thought prompting elicits explicit multi-step rationales from LLMs through guided instructions~\citep{kahneman2011thinking,sahoo2024systematic}. It has proven effective across diverse domains, including mathematics, scientific problem solving, and code generation, driving gains in accuracy and robustness~\citep{team2023gemini}. 
For instance, OpenAI’s o1 employs reinforcement learning to cultivate complex reasoning abilities, showing improvements under both training-time and test-time scaling~\citep{jaech2024openai}. Similarly, DeepSeek-R1~\citep{guo2025deepseek} and QwQ~\citep{qwq} demonstrate substantial benefits from CoT-based reasoning.

Notably, DeepSeek-R1 introduces GRPO, an RL framework that replaces learned critics with rule-based rewards, stabilizing post-training and enabling scaling to longer CoT. Extending CoT to the visual domain has also attracted increasing attention~\citep{team2025kimi,zhou2025r1,yang2025r1,wang2025vl,zheng2025deepeyes,peng2025skywork}. 
For example, Visual-RFT~\citep{liu2025visual} applies GRPO to detection, grounding, and classification tasks, while Vision-R1~\citep{huang2025vision} curates a large-scale image CoT dataset to train an R1-style visual reasoner.

Although CoT improves robustness on compositional and symbol-intensive tasks, it is not generally beneficial. Several studies report overthinking when tasks are primarily perceptual or intuitive~\citep{sui2025stop,kumar2025overthink,xie2025arm2,chen2024not}. Our analysis reveals a similar phenomenon in the video domain and motivates a reason-when-necessary strategy to mitigate unnecessary complexity and improve efficiency in video understanding.

\subsection{Video Reasoning Models}
Early work on video reasoning adapts R1-style reinforcement learning techniques from images to videos, such as Video-R1~\citep{feng2025video} and VideoChat-R1~\citep{li2025videochat}. Beyond QA, some approaches extend reasoning to temporal grounding tasks, \eg, Time-R1~\citep{wang2025time} shows that explicit reasoning can benefit temporal localization. Other efforts target specific designs such as relational reasoning over objects~\citep{gong2025reinforcing}, narrative reasoning across long videos~\citep{ghazanfari2025chain}, and scalable training~\citep{chen2025scaling,li2025veripo,fu2025love}.

Recent works further explore interleaved video-text reasoning, also known as ``thinking with frames''. These methods employ progressive perception strategies similar to ``thinking with images'' in the image domain, where the model first reasons to select salient frames or segments, then revisits them at higher resolution or frame rate to produce more accurate answers~\citep{zhang2025thinking,xie2025video}.

Despite these advances, prior methods enforce an \textit{always-thinking} paradigm for videos~\citep{wang2025videorft,feng2025video,chen2025exploring,zhang2025tinyllava,li2025veripo,luo2025museg,li2025reinforcement,park2025deepvideo}. Our analysis shows that on perception-oriented QA tasks, direct answering often matches CoT performance. This motivates a more adaptive approach: apply direct answering when it suffices and reserve CoT reasoning for cases where it yields tangible gains.

\subsection{Auto-Thinking}
To improve reasoning efficiency, auto-thinking methods aim to determine when to invoke CoT, typically by training a switching policy via SFT or RL~\citep{cheng2025incentivizing,lou2025adacot,kang2025c3ot,xie2025arm2,ma2025reasoning,sui2025stop,qu2025survey,chen2025aware,shen2503dast,chen2025towards,li2025system}. Among them, AdaptThink~\citep{zhang2025adaptthink} emphasizes the importance of balanced data sampling between think and no-think samples during on-policy training and achieves competitive performance on math tasks. In the image domain, R-4B~\citep{yang2025r} adopts bi-mode policy optimization, using SFT for initialization and then refining the model via RL to enhance the decision accuracy of whether to activate CoT. However, directly extending these strategies to video is non-trivial, as genuinely ``must-think’’ samples are relatively rare in videos~\citep{zhan2025kat}, which makes mode-switching supervision less stable during training.

Our \methodname{} departs from prior auto-thinking approaches in two aspects:  (1) During training, instead of supervising a binary mode for each sample, we train the model with both direct and CoT answers. This eliminates the need for think/no-think labels, switch tokens, or cold-start SFT. Empirically, this training strategy reduces mode collapse and improves generalization. (2) At inference, we compute the mean log probability of the first answer to determine whether to proceed with CoT, enabling controllable and efficient thinking-mode selection.

\section{Preliminaries}

In this section, we first briefly introduce our training framework and then analyze CoT inference versus direct inference in existing video reasoning models, revealing that indiscriminately enabling step-by-step reasoning is often redundant for video understanding.

\subsection{Training Framework}
\label{sec:grpo_framework}

\noindent\textbf{GRPO Training.} As a recent RL method, Group Relative Policy Optimization (GRPO) replaces a learned critic with group-normalized, rule-based verifiable rewards, offering a simplified and scalable RL training pipeline with strong empirical performance~\citep{guo2025deepseek}.

Formally, given a prompt $q$, the behavior policy $\pi_{\theta_{\text{old}}}$ samples $G$ candidate outputs $\{o_1,\dots,o_G\}$. For each output, a verifiable reward $r_i$, such as answer accuracy, temporal IoU, or format correctness, is computed. GRPO then normalizes these rewards using the group-wise mean $\mu$ and standard deviation $\sigma$ to obtain relative advantages $A_i = \frac{r_i - \mu}{\sigma+ \varepsilon}$. Then with the importance ratio $\rho_i = \frac{\pi_\theta(o_i \mid q)}{\pi_{\theta_{\text{old}}}(o_i \mid q)}$, the training objective becomes:
\begin{equation}
\begin{split}
\mathcal{L}_{\text{GRPO}}(\theta)
&= -\frac{1}{G}\sum_{i=1}^G
\min\!\Big(\rho_i A_i,\; \operatorname{clip}(\rho_i,\,1-\epsilon,\,1+\epsilon)\,A_i\Big)
+\beta\, D_{\text{KL}}\!\big(\pi_{\theta}\,\Vert\,\pi_{\text{ref}}\big)
\end{split}
\end{equation}
where $D_{\text{KL}}$ regularizes the policy against a reference policy $\pi_{\text{ref}}$ via a KL penalty, and $\beta \ge 0$ controls the strength of this regularization.

\noindent\textbf{Reward Function.} Standard GRPO employs verifiable, rule-based rewards consisting of a task-accuracy term $R_{\text{task}}$ and a format correctness term $R_{\text{fmt}}$. The final per-sample reward is defined as a weighted sum:
\[
R_i \;=\; w\,R_{\text{task}}(o_i) \;+\; \lambda\,R_{\text{fmt}}(o_i),
\qquad w,\lambda\ge 0.
\]
In this paper, we consider three video task types: QA, temporal grounding, and grounding QA. The detailed reward for each task can be found in Appendix \ref{appendix:reward_design}.

\noindent\textbf{Training Data.} 
While traditional video reasoning models are trained primarily on videos, raw video data is inherently noisy and non-symbolic, often biasing models toward perception rather than reasoning. To enhance the model's long-chain reasoning capabilities, we augment the training corpus with high-quality text~\citep{yu2025dapo} and image sources~\citep{wang2025vl,wang2025sota} that cover math and scientific problems. We also include video QA data~\citep{feng2025video,cores2024tvbench,li2025sti,zhu2025mmr} and temporal grounding data~\citep{gao2017tall,caba2015activitynet,wang2025time,xiao2024can}. After filtering, we obtain 83K samples. The detailed training data can be found in Appendix \ref{appendix:data}.

\textbf{Direct RL without Cold-Start.} Notably, we conduct RL directly on the curated data without relying on a cold-start SFT stage. Collecting large-scale, high-quality multimodal CoT traces is expensive and often noisy. In early experiments, SFT on Video-R1-CoT data~\citep{feng2025video}, which has both the intermediate reasoning traces and final answer, degraded the Qwen2.5-VL baseline~\citep{bai2025qwen2}. We therefore focus on directly incentivizing the base model’s reasoning via reinforcement learning. The detailed ablations can be found in Appendix~\ref{cold_start_ablation}. 

\subsection{Analysis of Existing Video Reasoning Models}

\begin{table*}[t]
    \centering
    \footnotesize
    \setlength{\tabcolsep}{0.91em}
    \caption{\textbf{Comparison of Direct and CoT Inference for Video Reasoning Models.} Direct inference means answering without explanations. CoT inference follows each model's default prompt to elicit step-by-step reasoning and then generate the final answer. All models are re-evaluated with the same inputs, \ie, maximum 256 frames and 16K total video tokens. We report the accuracy and the response length (in tokens). Surprisingly, CoT inference shows worse accuracy than direct inference while using more tokens on several benchmarks.}
    \begin{tabular}{lcclllll}
    \toprule
    \multirow{2}{*}{\textbf{Model}}     & \multirow{2}{*}{\textbf{\makecell{Inference \\ Strategy}}} & \multirow{2}{*}{\textbf{\makecell{Response \\ Length}}}   & \multirow{2}{*}{\textbf{VideoMME}}             & \multirow{2}{*}{\textbf{LongVideoBench}}                 & \multirow{2}{*}{\textbf{MMVU}}      & \multirow{2}{*}{\textbf{VideoMMMU}}            & \multirow{2}{*}{\textbf{Charades-STA}}  \\
    & \\
    \midrule
     Qwen2.5-VL & Direct  & 10.2 & 66.0 & 60.9 & 65.7 & 52.7  & 52.9 \\
    \midrule
    \multirow{2}{*}{Video-R1}   & Direct     & 17.6 & 64.6 & 59.5 & 65.6 & 51.4  & 42.0\\
               & CoT & \textcolor{blue}{386} & 64.3\textcolor{red}{($-0.3$)} & 59.4\textcolor{red}{($-0.1$)} & 65.4\textcolor{red}{($-0.2$)} & 52.4\textcolor{OliveGreen}{($+1.0$)}  & 34.9\textcolor{red}{($-7.1$)} \\
    \midrule
    \multirow{2}{*}{Time-R1}   & Direct & 9.2 & 65.9 & 60.0 & 65.1 & 53.0  & 56.6 \\
    & CoT & \textcolor{blue}{138} & 63.8\textcolor{red}{($-2.1$)} & 58.3\textcolor{red}{($-1.7$)} & 64.7\textcolor{red}{($-0.4$)} & 54.1\textcolor{OliveGreen}{($+1.1$)}  & 58.8\textcolor{OliveGreen}{($+2.2$)}\\
    \midrule
    \multirow{2}{*}{VideoChat-R1}  & Direct & 4.3 & 65.7 & 60.1 & 65.6 & 52.3  & 58.5 \\
    & CoT & \textcolor{blue}{126} & 63.9\textcolor{red}{($-1.8$)} & 58.2\textcolor{red}{($-1.9$)} & 65.4\textcolor{red}{($-0.2$)} & 55.7\textcolor{OliveGreen}{($+3.4$)}   & 59.9\textcolor{OliveGreen}{($+1.4$)} \\
    \bottomrule
    \end{tabular}
\label{table:think_direct_analysis}
\end{table*}

Before building our own reasoning model, we pose the following question: 

\begin{center}
\begin{tcolorbox}[width=1\textwidth,coltext=metafg,colback=metabg]
\vspace{-1mm}
\textit{When is video chain-of-thought actually necessary, and how does it compare with direct answering?}
\vspace{-1mm}
\end{tcolorbox}
\end{center}

To investigate, we re-evaluate existing video reasoning models, \ie, Video-R1~\citep{feng2025video}, Time-R1~\citep{wang2025time}, and VideoChat-R1~\citep{li2025videochat}, which are all based on Qwen2.5-VL. We compare two inference strategies: direct inference and CoT inference. Results are summarized in Table~\ref{table:think_direct_analysis}.

Surprisingly, direct inference often matches, or even outperforms, CoT inference on several benchmarks such as VideoMME~\citep{fu2025video} and LongVideoBench~\citep{wu2024longvideobench}, while generating significantly fewer tokens (see Figure~\ref{fig:examples_videochat_fail}). Consistent CoT gains are primarily observed on Video-MMMU~\citep{hu2025video}. We further examine the samples where CoT succeeds but direct inference fails (see Figure~\ref{fig:examples_videochat_fail_success}). These cases are typically math- or physics-oriented (\eg, physics instructional videos with blackboard derivations): the questions or answer options contain symbolic inputs, the visual signal is relatively clean, and multi-step deduction is genuinely necessary. Under these conditions, CoT provides a tangible advantage.

By contrast, in perception-oriented queries (\eg, object or action recognition, simple attribute identification), CoT often redundantly describes the video or compares answer options step by step, yet ultimately arrives at the same conclusion as direct inference. Given the autoregressive nature of LLMs, such verbose traces substantially increase end-to-end latency and inference cost. Considering that most QA samples do not benefit from additional reasoning, we believe an effective and efficient policy is to \emph{reason only when necessary}, that is, employ \textbf{auto-thinking}. Accordingly, in this paper, we focus on building an auto-thinking video model.

\section{\methodname{}}

\begin{figure*}[t]
\centering
\includegraphics[width=0.99\linewidth]{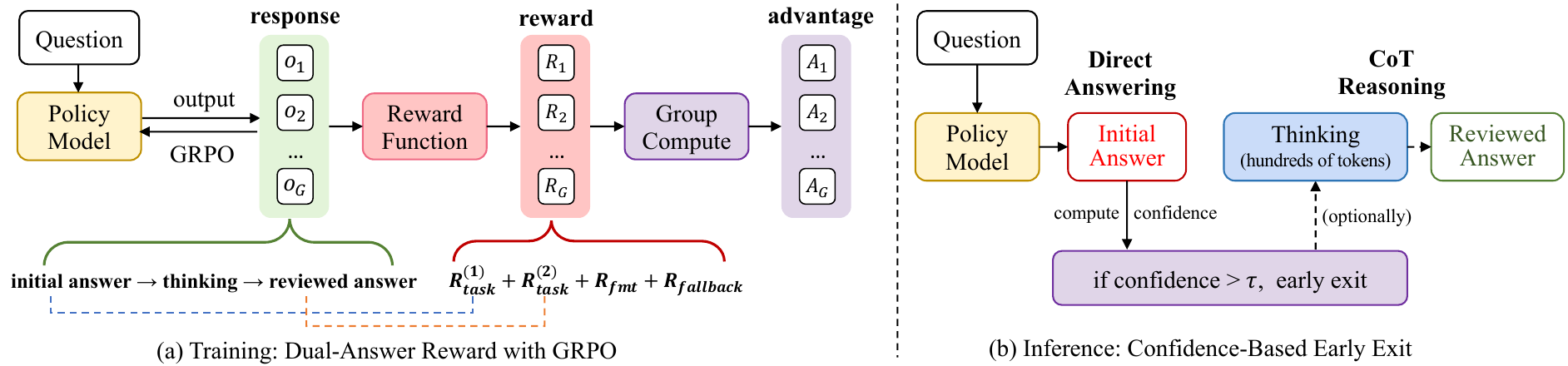}
\caption{\textbf{Overview of \methodname{}.} (a) \textbf{Training:} The response follows the \emph{answer $\rightarrow$ think $\rightarrow$ answer} template, jointly optimizing both the initial and reviewed answers. Specifically, a fallback reward is introduced to avoid a spurious initial guess. (b) \textbf{Inference:} The model first produces an initial answer. If its length-normalized confidence exceeds a threshold $\tau$, decoding terminates as direct answering; otherwise, the model continues with CoT reasoning and outputs a reviewed answer, enabling adaptive, confidence-based early exit.}
\label{fig:method}
\end{figure*}

In this section, we present \textbf{\methodname{}}, a simple yet effective framework that reasons only when necessary, as illustrated in Figure~\ref{fig:method}. During training, we adopt an \emph{answer $\rightarrow$ think $\rightarrow$ answer} template.
At inference time, an early-exit mechanism determines whether to continue reasoning after the first answer.

\subsection{Thinking Once, Answering Twice}

A common approach to auto-thinking involves learning a mode-switching policy during training, \eg, randomly dropping CoT traces in SFT so the model alternates between direct and CoT outputs~\citep{zhang2025adaptthink}. While effective on text, it depends on careful data balancing and is sensitive to training hyperparameters. In video, the scarcity of high-quality reasoning examples further exacerbates instability.

We adopt a different perspective: genuine CoT should be built on top of an initial answer. For easy questions, the initial answer should suffice; for harder ones, the model should verify and revise its response within the same generation. Accordingly, we do not train separate “think” and “direct” modes. Instead, the model always learns to generate a concise first answer and a reasoned second answer. This design avoids the need for per-sample mode labels, specialized switch tokens or heads, or other artifacts. The distinction between direct and thinking modes is made solely at test time through a confidence-based early-exit mechanism.

\noindent\textbf{Output Format.} Given a prompt $q$, each training response $o$ follows a strict, verifiable format:
\begin{center}
{\color{blue}\verb|\\boxed{|$a_1$\verb|}|} {\color{RedOrange}\verb|<think>|$r$\verb|</think>|}{\color{blue} \verb|\\boxed{|$a_2$\verb|}|}
\end{center}
Here, $a_1$ and $a_2$ are short, verifiable answers, and $r$ is a free-form rationale. We enforce exactly two \verb|\\boxed{...}| blocks and one \verb|<think>...</think>| block, with no extra text before/after. To achieve such an output format, a system prompt (Table~\ref{table:visionthink_prompt}) is carefully designed, enabling generation without cold-start SFT.

\noindent\textbf{Fallback Tolerance.}
For mathematically or symbolically complex problems, the model may be unable to produce a correct $a_1$ without intermediate reasoning~\citep{yue2025don}. To prevent low-confidence guesses, we provide a designated fallback string. When immediate answering is infeasible, the model outputs \textit{``Let's analyze the problem step by step''} in the first box, then proceeds to reasoning and produces the final answer $a_2$. This design preserves the output grammar, avoids spurious guesses, and ensures the early-exit mechanism remains unambiguous and interpretable.

\noindent\textbf{Why ``answer–think–answer''?}
This template decouples \emph{when} to think, handled at test time by our early-exit rule, from \emph{how} to think, namely the reasoning behavior learned during RL training. Empirically, this design yields more stable training for videos with less data effort than traditional mode-switching approaches~\citep{zhang2025adaptthink}. It also makes inference easy to control: with ample compute, one can always use the reviewed answer, while under tight budgets the model can fall back to the initial direct answer and still benefit from RL training. Overall, this decoupling of the training objective and inference policy gives users flexible control over the trade-off between accuracy and efficiency.

\begin{table}[t]
    \centering
    \footnotesize
    \caption{\textbf{System Prompt for \methodname{}.} The prompt follows an \emph{answer $\rightarrow$ think $\rightarrow$ answer} template, enabling both direct and CoT outputs in one generation.} 
    \vspace{-4pt}
        \begin{tabular}{p{16cm}}
        \toprule
        \textbf{\textit{SYSTEM PROMPT}} \\
        \midrule
        You are a helpful assistant. \\
        FIRST: Output your initial answer inside the first {\color{blue}\verb|\\boxed{...}|} without any analysis or explanations. If you cannot determine the answer without reasoning, output {\color{blue}\verb|\\boxed{Let's analyze the problem step by step.}|} instead.\\
        THEN: Think through the reasoning as an internal monologue enclosed within {\color{RedOrange}\verb|<think>...</think>|}. \\
        AT LAST: Output the final answer again inside {\color{blue}\verb|\\boxed{...}|}. If you believe the previous answer was correct, repeat it; otherwise, correct it. \\
        Output format: {\color{blue}\verb|\\boxed{...}|}{\color{RedOrange}\verb|<think>...</think>|}{\color{blue}\verb|\\boxed{...}|} \\
        \bottomrule
    \end{tabular}
\label{table:visionthink_prompt}
\vspace{-6pt}
\end{table}

\subsection{Training: Dual-Answer Reward with GRPO}

We follow the GRPO framework described in Section~\ref{sec:grpo_framework}, but introduce a new \emph{dual-answer} reward that supervises both the initial and reviewed answers. Let $a_1$ and $a_2$ denote the first and second boxed answers, respectively. The total reward is given by:
\begin{equation*}
\label{eq:total-reward}
R\;\!=\!\;
w_1\,R_{\text{task}}^{(1)}(a_1)
\;\!+\!\;
w_2\,R_{\text{task}}^{(2)}(a_2)
\;\!+\!\;
\lambda\,R_{\text{fmt}}
\;\!+\!\;
\alpha\,R_{\text{fallback}}
\end{equation*}
where $w_2 > w_1 \ge 0$, and $\lambda, \alpha \ge 0$ are weight coefficients. 

The task rewards $R_{\text{task}}$ follow the previous definitions. Notably, we assign a higher weight $w_2$ to the final answer $a_2$ to encourage more accurate reviewed responses while still incentivizing good initial answers. This design also penalizes cases where the first answer is correct but the second is incorrect, pushing the model to improve overall reliability. The term $R_{\text{fmt}}$ ensures that the output format adheres to the required \emph{answer $\rightarrow$ think $\rightarrow$ answer} template.

Particularly, the last term $R_{\text{fallback}} \in \{0,1\}$ is a \emph{fallback bonus} when $a_1$ is the designated string \textit{``Let's analyze the problem step by step''} and $a_2$ is correct. This discourages low-confidence guesses in $a_1$ for difficult problems and rewards honest deferral followed by accurate reasoning. It is particularly helpful for math and symbol-heavy questions, where premature guesses are often wrong. Further analysis of the reward design is discussed in Appendix~\ref{appendix:reward_analysis}.

During training, we observe consistent increases in total reward. Notably, $R_{\text{task}}^{(2)}$ typically exceeds $R_{\text{task}}^{(1)}$, confirming the benefit of explicit reasoning for more challenging instances while still retaining fast, correct first answers when appropriate. 

\subsection{Inference: Confidence-Based Early Exit}
\label{subsec:inference}

To enable adaptive and controllable reasoning, we adopt a simple yet effective \emph{early-exit} mechanism, where a rule-based check determines whether the first boxed answer has sufficient confidence to justify skipping the rest of the generation. Prior study~\citep {liao2025fractured} also shows that token-level confidence correlates strongly with answer correctness in modern LLMs. We leverage this finding to score the model’s own output directly, without relying on external calibrators.

At inference, we first decode only up to the closing delimiter of the first boxed answer. Let $a_1 = (t_1, \dots, t_L)$ denote the tokens within the first box. We compute the following length-normalized confidence score:
\begin{equation}
s(a_1) \;=\; \frac{1}{L}\sum_{\ell=1}^{L}\log p_\theta\!\left(t_\ell \mid t_{<\ell},\, q\right),
\label{eq:score}
\end{equation}
where $p_\theta$ is the model’s next-token distribution under the chosen decoding policy. If $a_1$ is the fallback string, we set $s(a_1) = -\infty$, forcing continuation of the CoT and final answer generation.

Given a confidence threshold $\tau$, we accept $a_1$ and terminate decoding if $s(a_1) \ge \log \tau$; otherwise, we continue to generate the rationale $r$ and second answer $a_2$. The threshold $\tau$ controls the accuracy–efficiency trade-off and can be determined on a held-out set. In practice, a single fixed threshold works well across diverse video QA benchmarks. Besides, since $a_1$ typically consists of fewer than ten tokens, the confidence score is fast to compute. In many cases, early exit avoids generating hundreds of additional tokens, substantially reducing latency and inference cost. 

\section{Experiments}

\subsection{Experiment Details} 

\textbf{Implementation Details.}
Our models are fine-tuned from Qwen2.5-VL-7B-Instruct~\citep{bai2025qwen2} and Qwen3-VL-8B-Instruct~\citep{bai2025qwen3}. During training, the maximum number of total video tokens is set to 4{,}096, and the maximum number of frames is set to 256. We use AdamW as the optimizer, with a learning rate of $1\!\times\!10^{-6}$, weight decay of $0.01$, and a maximum gradient norm of $1.0$. A constant learning rate schedule without warm-up is employed. The KL penalty coefficient $\beta$ is set to $0.01$. Task reward weight $w_{1}$ is $0.9$ and $w_{2}$ is $1.1$; the format reward weight $\lambda_{\text{fmt}}$ is $1$; and the fallback reward weight $\alpha$ is $0.3$. The global batch size is set to 256, and we train the model for one epoch. The visual encoder remains frozen; only the projector and the LLM are fine-tuned. We leverage DeepSpeed~\citep{rasley2020deepspeed} and vLLM~\citep{kwon2023efficient} to accelerate the training. For GRPO rollout generation, we set the rollout size $G$ to 16 and use a temperature of $1.0$ to encourage exploration. Our training is conducted on 32 H100 GPUs for approximately 35 hours.

During testing, all evaluations are performed using \texttt{lmms-eval} \citep{zhang2024lmmsevalrealitycheckevaluation} with greedy decoding (temperature 0). The maximum response length is set to 4,096 tokens, ensuring that no truncation occurs during evaluation. The early-exit confidence threshold $\tau$ is set to $0.97$, which is tuned on a small held-out subset of the validation data and generalizes well to unseen evaluation benchmarks. For the Qwen2.5-VL model, we allow up to 16K total video tokens and vary the maximum number of frames among \{64, 128, 256\}. For the Qwen3-VL model, we allow up to 128K total video tokens and sweep over \{64, 256, 2048\} frames. Following \cite{bai2025qwen2} and \cite{bai2025qwen3}, we report the highest performance across these settings.

\noindent\textbf{Evaluation Benchmarks.} 
We evaluate on both video QA and temporal grounding benchmarks. For perception-oriented QA, we report accuracy on VideoMME (without subtitles) \citep{fu2025video}, MVBench \citep{li2024mvbench}, LongVideoBench \citep{wu2024longvideobench}, and MMVU (multi-choice) \citep{zhao2025mmvu}.
To assess reasoning-intensive tasks, we evaluate on VideoMMMU~\citep{hu2025video} and Minimal Video Pairs (MVP)~\citep{krojer2025shortcut}.  Particularly, for MVP, visually similar videos are paired with identical questions but opposing answers. Models must answer both correctly, and we report pairwise accuracy on the MVP-mini subset. For temporal grounding, we report recall and mean IoU on Charades-STA~\citep{gao2017tall} and ActivityNet~\citep{caba2015activitynet}. Finally, we use NExT-GQA~\citep{xiao2024can} to evaluate grounding QA performance.

Additionally, we evaluate our model on image reasoning benchmarks, such as MathVista~\citep{lu2023mathvista}, MathVision~\citep{wang2024measuring}, MathVerse~\citep{zhang2024mathverse}, MMMU~\citep{yue2024mmmu}, MMMU-Pro~\citep{yue2025mmmu}, and MM-Vet~\citep{yu2023mm}.

\subsection{Main Results}

\begin{table*}[t]
    \centering
    \footnotesize
    \setlength{\tabcolsep}{0.13em}  
    \caption{\textbf{Evaluation Results on Video QA Benchmarks.} We compare \methodname{} with thinking-only video reasoning models on both perception-oriented and reasoning-heavy benchmarks, and also report the average response length (in tokens). $\dagger$ means reproduced results.
    We also report the think ratio, defined as the proportion of samples on which the model triggers CoT reasoning.
    Relative to the Qwen baseline, \methodname{} yields consistent and more pronounced gains on the reasoning benchmarks. We further observe that the thinking ratio is low on perception-oriented benchmarks but substantially higher on reasoning-heavy ones. }  
    \begin{tabular}{lccccccccc}
    \toprule
    \multirow{3}{*}{\textbf{Model}}                     & \multirow{3}{*}{\textbf{\makecell{Reasoning \\ Mode}}}      &   \multirow{3}{*}{\textbf{\makecell{Response \\ Length}}} & \multicolumn{4}{c}{\textbf{Video Perception Benchmark}} & & \multicolumn{2}{c}{\textbf{Video Reasoning Benchmark}} \\
    \cmidrule{4-7}\cmidrule{9-10}
                            &    &               & VideoMME             & MVBench             & LongVideoBench    & MMVU     &     & VideoMMMU             & MVP  \\
    \midrule
    Qwen2.5-VL-7B$^{\dagger}$ & \xmark & 3.0 & 66.0 & 67.1 & 60.9 & 66.2 & & 54.7  & 36.5 \\
    Qwen3-VL-8B$^{\dagger}$ & \xmark & 2.2 & 72.5 & 69.4 & 67.6 & 69.9 & & 61.0  & 40.5 \\
    \midrule
    Temporal-RLT    & Think-Only & - & 57.6 & 68.1 & - & 65.0  & & -& - \\
    Video-RFT        & Think-Only & -  & 59.8 & 62.1 & - & 68.5& & 51.1  & - \\
    Video-R1         & Think-Only & 386 & 61.8 & 65.5 & - & 65.0 & & 51.4  & 33.0 \\
    Video-RTS       & Think-Only & - & 63.0 & - & 56.6 & 66.4 & & 52.7  & - \\
    VITAL  & Think-Only & - & 64.1 & -  & - & 68.7 & & 54.2  & - \\
    LongVILA-R1      & Think-Only & - & 65.1 & 67.6  & 58.0 & -& &-  & -  \\
    LOVE-R1         & Think-Only & - & 66.2 & 66.6 & 60.1 & -& & -  & - \\
    VideoChat-R1.5   & Think-Only & 133 & 65.2 & 70.6 & 61.4 & -& & 49.6  & 38.6 \\
    \midrule
    \rowcolor{metabg}
    \textbf{\methodname{}} \tiny{(Qwen2.5-VL-7B)} & \textbf{AutoThink} & \textbf{44} & \textbf{67.3} & \textbf{71.0} & \textbf{60.5} & \textbf{69.7} & & \textbf{58.6}  & \textbf{39.4} \\
    (think ratio)  & &  & (40\%) & (25\%) & (39\%) & (28\%)& & (51\%)  & (44\%) \\
    \rowcolor{metabg}
    \textbf{\methodname{}} \tiny{(Qwen3-VL-8B)} & \textbf{AutoThink} & \textbf{52} & \textbf{71.7} & \textbf{72.0} & \textbf{67.4} & \textbf{71.1} & & \textbf{65.0}  & \textbf{43.0} \\
    (think ratio) & & & (11\%) & (31\%) & (20\%) & (38\%) & & (53\%) & (56\%) \\
    \bottomrule
    \end{tabular}
\label{table:benchmark_qa}
\end{table*}

\textbf{Video QA Benchmarks.}
As shown in Table~\ref{table:benchmark_qa}, our \methodname{} achieves state-of-the-art results on both perception and reasoning benchmarks. Concretely, \methodname{} achieves 67.3\% accuracy on VideoMME with a Qwen2.5-VL base, surpassing previous reasoning models such as Video-R1~\citep{feng2025video}, VITAL~\citep{zhang2025thinking}, and VideoChat-R1.5~\citep{yan2025videochat} by 5.5\%, 3.2\%, and 2.1\% respectively. On the reasoning-intensive VideoMMMU benchmark, \methodname{} improves accuracy from 54.7\% to 58.6\% (+3.9\%), and on the harder MVP benchmark, it increases pairwise accuracy from 36.5\% to 39.4\%, consistently outperforming existing reasoning models such as Video-R1 by a large margin of $\sim$6\% accuracy.
When built on Qwen3-VL, our \methodname{} further improves performance and achieves a remarkable 65.0\% on VideoMMMU. These results demonstrate that our auto-thinking is effective for video understanding.

Beyond accuracy, \methodname{} also substantially improves inference efficiency. Compared to Video-R1’s 386-token responses, our model generates only 44 tokens on average. Moreover, the model adaptively triggers reasoning depending on task complexity: the think-mode activation ratio is only 25\% on perception-oriented MVBench, while it rises to 51\% on the reasoning-heavy Video-MMMU. This indicates that our model can invoke CoT for genuinely challenging queries, highlighting the inference efficiency of our auto-thinking. 

\noindent\textbf{Temporal Grounding Benchmarks.}  Results on temporal grounding benchmarks are summarized in Table~\ref{table:benchmark_grounding}. Notably, after dual-answer GRPO training, the initial boxed prediction is already sufficient for accurate localization. The subsequent CoT trace mainly provides explanatory interpretation without improving localization performance. We therefore adopt early exit by default to improve inference efficiency. More discussion can be found in Appendix~\ref{appendix:analysis_on_grounding}. 

Overall, \methodname{} improves mIoU from 52.9\% to 60.0\% on Charades-STA and from 26.9\% to 47.6\% on ActivityNet, surpassing Time-R1~\citep{wang2025time} and VideoChat-R1.5~\citep{yan2025videochat}. On NExT-GQA, QA accuracy is also improved from 53.3\% to 80.6\%, and mIoU is improved from 20.2\% to 36.7\%. With Qwen3-VL, all grounding metrics further increase, setting new state-of-the-art results. These experiments validate the effectiveness of our models for temporal grounding. 

\begin{table*}[t]
    \centering
    \footnotesize
    \setlength{\tabcolsep}{0.85em}  
    \caption{\textbf{Evaluation Results on Temporal Grounding Benchmarks.} $\dagger$ means reproduced results. We observe that on grounding benchmark, the initial boxed answer is sufficient, so we early-exit without further reasoning to save computation.}
    \begin{tabular}{lcccccccccccc}
    \toprule
    \multirow{3}{*}{\textbf{Model}}                         & \multicolumn{12}{c}{\textbf{Temporal Grounding Benchmark}} \\
    \cmidrule{2-13}
                                        &  \multicolumn{4}{c}{Charades-STA} & & \multicolumn{4}{c}{ActivityNet} & & \multicolumn{2}{c}{NExT-GQA} \\
    & 0.3 & 0.5 & 0.7 & mIoU & & 0.3 & 0.5 & 0.7 & mIoU & & Acc & mIoU\\
    \midrule
    Qwen2.5-VL-7B$^\dagger$ & 77.7 & 59.6 & 34.8 & 52.9 & & 37.9 & 22.6 & 10.6 & 26.9 & & 53.3 & 20.2 \\
    \midrule
    TimeChat & 51.0 & 27.5 & 11.4 & 31.2 & & 44.0 & 27.8 & 14.3 & 30.4 & & 28.8 & 17.4 \\
    TimeSuite & 79.4 & 67.1 & 43.0 & - &  & - & -& - & - & & -& -\\
    TimeMarker & 73.5 & 51.9 & 26.9 & 48.4 & & 67.4 & 50.7 & 33.0 & 49.5 & & - & - \\
    Temporal-RLT    & 79.6 & 67.9 & 44.1 & 57.0 & & 56.9 & 38.4 & 20.2 & 39.0 & & 78.7 & 37.3 \\
    Time-R1   & 82.8 & 72.2 & 50.1 & 58.8 & & 73.3 & 55.6 & 34.0 & 52.1 & & - & - \\
    VITAL & 83.1 & 72.0 & 46.7 & 59.9 & & 70.9 & 50.8 & 31.6 & 49.8 & & 78.7 & 43.0 \\
    VideoChat-R1.5 & 82.8 & 71.6 & 48.3 & 60.6 & & 52.4 & 32.3 & 16.8 & 35.3 & & - & - \\
    \midrule
    \rowcolor{metabg}
    \textbf{\methodname{}} \tiny{(Qwen2.5-VL-7B)} & \textbf{82.9} & \textbf{70.8} & \textbf{46.0} & \textbf{60.0} & & \textbf{69.2} & \textbf{48.5} & \textbf{27.3} & \textbf{47.6} & & \textbf{80.6} & \textbf{36.7} \\
    \rowcolor{metabg}
    \textbf{\methodname{}} \tiny{(Qwen3-VL-8B)} &  \textbf{85.1} & \textbf{74.9} & \textbf{53.7} & \textbf{63.7} & & \textbf{74.1} & \textbf{54.3} & \textbf{32.4} & \textbf{51.9} & & \textbf{81.1} & \textbf{44.2} \\
    \bottomrule
    \end{tabular}
\label{table:benchmark_grounding}
\end{table*}

\noindent\textbf{Image Understanding Benchmarks.}
Although \methodname{} is primarily designed for video understanding, we also evaluate its performance on several image reasoning benchmarks. As shown in Table~\ref{table:benchmark_image}, \methodname{} consistently outperforms the Qwen baseline. These improvements are largely attributable to the inclusion of image-centric math and reasoning data during training, which strengthens the model’s visual reasoning skills beyond the video domain. At the same time, the thinking once, answering twice design and dual-answer reward transfer naturally to static images, where the model can still benefit from an internal reasoning stage before giving a reviewed answer. Together, the results demonstrate that \methodname{} is not only effective for video understanding, but also exhibits strong generalization to challenging image benchmarks.

\begin{table*}[t]
    \centering
    \footnotesize
    \caption{\textbf{Evaluation Results on Image Benchmarks.} The Qwen and \methodname{} are evaluated under the same settings.}
    \begin{tabular}{lcccccc}
    \toprule
    \textbf{Model}           & \makecell{\textbf{MathVista}\\testmini}  &  \makecell{\textbf{MathVision}\\testmini} & \makecell{\textbf{MathVerse}\\testmini} &  \makecell{\textbf{MMMU}\\val} &  \makecell{\textbf{MMMU-Pro}\\overall} &  \makecell{\textbf{MM-Vet}\\test}\\
    \midrule
    Qwen2.5-VL-7B & 69.4 & 26.3 & 44.8 & 51.3 & 36.1 & 60.0 \\
    \textbf{\methodname{}}\tiny{(Qwen2.5-VL-7B)} & \textbf{73.7} & \textbf{29.6}  & \textbf{46.9} & \textbf{53.8} & \textbf{39.8} & \textbf{61.9} \\
    \bottomrule
    \end{tabular}
\label{table:benchmark_image}
\end{table*}

\subsection{Analyses and Ablations}

To verify the effectiveness of different design choices, we conduct the following analyses and ablations. Unless specified, all experiments use models built on Qwen2.5-VL.

\noindent{\textbf{Comparison between Different Training Strategies.} 
To clearly demonstrate the advantages of RL and auto-thinking, we compare four training strategies on the same data: (1) SFT, which directly predicts an answer; (2) RL without thinking, which applies GRPO on direct answers without prefix reasoning; (3) RL with thinking, which generates CoT then an answer optimized with standard GRPO; and (4) our auto-thinking, which adaptively chooses direct/CoT. The results are summarized in Table~\ref{table:ablation_training_strategy}, and the prompt templates used are provided in Appendix~\ref{appendix:prompt_template}.

Several key observations emerge. First, direct answering methods (SFT and RL without thinking) bring only mild gains over the baseline; RL without thinking performs better on format-sensitive tasks such as Charades-STA, indicating improved robustness. Second, RL with thinking substantially boosts reasoning-heavy benchmarks like VideoMMMU, but inflates the average response length from 2.5 to 149 tokens and offers limited benefits on perception-oriented tasks such as VideoMME and MMVU, suggesting that chain-of-thought reasoning is redundant for simpler tasks. In contrast, \methodname{} outperforms all variants (\eg, +3.9\% on VideoMMMU, +1.3\% on VideoMME) while cutting the average response length to 44 tokens by invoking CoT only when necessary. In particular, compared with RL with thinking, \methodname{} achieves higher accuracy since both the initial and reviewed answers are explicitly supervised and jointly optimized during reinforcement learning.

\begin{table*}[t]
    \centering
    \footnotesize
    \caption{\textbf{Comparison between Different Training Strategies.} 
    \methodname{} delivers stronger performance than SFT on reasoning and grounding benchmarks, and surpasses standard RL with CoT while reducing response length, achieving efficiency and efficacy.}
    \begin{tabular}{lccllll}
    \toprule
    \textbf{Training}     & \textbf{Inference}   & \textbf{Response Length} & \textbf{VideoMME}           & \textbf{MMVU}        & \textbf{VideoMMMU}            & \textbf{Charades-STA}  \\
    \midrule
    Qwen2.5-VL-7B & Direct & 3.0 & 66.0  & 66.2 & 54.7  & 52.9 \\ 
    \midrule
    SFT & Direct & 2.3 & 67.0  & 65.9   & 56.5  & 56.3 \\
    \midrule
    RL without Thinking & Direct & 2.5 & 66.0  & 66.4 & 54.4  & 58.8 \\
    RL with Thinking & CoT & 149 & 66.1 & 67.5 & 56.4  & 59.8 \\
    \midrule
    \rowcolor{metabg}
    \textbf{\methodname{}} & \textbf{Direct/CoT} & \textbf{44} & \textbf{67.3}\textcolor{OliveGreen}{($+1.3$)} & \textbf{69.7}\textcolor{OliveGreen}{($+3.5$)} & \textbf{58.6}\textcolor{OliveGreen}{($+3.9$)}  & \textbf{60.0}\textcolor{OliveGreen}{($+7.1$)}\\
    \bottomrule
    \end{tabular}
\label{table:ablation_training_strategy}
\end{table*}

\noindent\textbf{Comparison with Other Adaptive Reasoning Strategies.} 
To further evaluate the effectiveness of our inference-based thinking-mode selection, we compare \methodname{} with a training-based strategy inspired by AdaptThink~\citep{zhang2025adaptthink}. In the training-based variant, each sample is labeled as think or no-think by comparing the average accuracy over 8 rollouts, and the model is then trained to either output a direct answer or produce a CoT followed by a final answer. To avoid collapse into a single mode, we maintain the think/no-think ratio close to 1:1. 

However, this approach brings limited gains. As shown in Table~\ref{table:autothink_ablation}, the auto mode of the training-based variant underperforms the no-think baseline on MVBench (70.5\% vs 71.1\%), and it behaves more similarly to no-think on reasoning benchmarks. It also suffers from mode collapse, defaulting to almost no thinking on VideoMME, and only a 31\% think ratio on VideoMMMU.

In contrast, \methodname{} applies confidence-based early-exit at inference time. It typically surpasses the no-think baseline, approaches the accuracy of always-think with much shorter responses, and consistently outperforms the training-based auto-thinking variant without extra labels or balancing, indicating that inference-time selection is stable and effective for adaptive video reasoning.

\begin{table*}[t]
    \centering
    \footnotesize
    \setlength{\tabcolsep}{0.55em}  
    \caption{\textbf{Comparison with Other Adaptive Reasoning Strategies.} 
    For comparison, we reproduce a \emph{training-based} auto-thinking baseline following~\cite{zhang2025adaptthink} that assigns think labels during RL. Results show that our \textit{inference-based} selection yields higher and more stable accuracy across benchmarks. In contrast, the training-based approach can even underperform direct answering. 
    }
    \begin{tabular}{lccccccc}
    \toprule
    \textbf{Inference Setting}   & \textbf{\makecell{Think Ratio}} & \textbf{\makecell{Response Length}} & \textbf{VideoMME} & \textbf{MVBench} & \textbf{MMVU} & \textbf{VideoMMMU}  & \textbf{MVP} \\
    \midrule
    \multicolumn{4}{l}{\textit{Training-Based Thinking-Mode Selection~\citep{zhang2025adaptthink}}} \\
    No-Think & 0\% & 23 & 67.1 & 71.1 & 65.3 & 55.4 & 36.5 \\
    Always-Think & 100\% & 166 & 66.3 & 70.3 & 68.0 & 54.8 & 39.3 \\
    \textbf{Auto}     &  \textbf{14\%}  & 31 & \textbf{67.1} & \textbf{70.5} & \textbf{67.2} & \textbf{55.7} & \textbf{36.8} \\
    (think ratio)  & &  & (1\%) & (0\%)  & (9\%) & (31\%) & (18\%) \\
    \midrule
    \multicolumn{4}{l}{\textit{Inference-Based Thinking-Mode Selection (Ours)}} \\
    Use 1st answer (\textit{no-think}) & 0\% & 8 & 67.3 & 70.9 & 69.3 & 54.6 & 39.0 \\
    Use 2nd answer (\textit{always-think}) & 100\% & 91 &67.3 & 71.0 & 69.8 & 58.7 & 39.8  \\
    \rowcolor{metabg}
    \textbf{\methodname{}} & \textbf{41\%} & \textbf{44} & \textbf{67.3} & \textbf{71.0} & \textbf{69.7} & \textbf{58.6} & \textbf{39.4} \\
    (think ratio)  & & & (40\%) & (25\%) & (39\%) & (51\%) & (44\%) \\
    \bottomrule
    \end{tabular}
\label{table:autothink_ablation}
\end{table*}

\noindent\textbf{Analysis of Confidence-Based Early Exit Mechanism.} 
In our inference strategy, we employ a confidence-based early exit mechanism to decide whether to invoke CoT reasoning after the initial answer. We hypothesize that the model’s token-level confidence in the first answer correlates with the need for further reasoning, which we empirically validate in Table~\ref{table:dataset_prob}. 

On perception-oriented benchmarks such as MVBench and MMVU, the average confidence of the initial answer (mean probability) exceeds 0.93, the think ratio remains around 25\%, and CoT yields only marginal gains, indicating that direct answers are sufficient. In contrast, on the reasoning-heavy benchmark VideoMMMU, average confidence drops to approximately 0.87, the think ratio increases to 51\%, and we observe a clear 4\% accuracy gain, showing that the mechanism successfully allocates more reasoning budget to harder tasks where CoT provides a tangible advantage.

We further examine whether this confidence signal captures truly \textit{think-needed} cases by measuring the recall of the predicted thinking mode on samples where $a_1$ is wrong but $a_2$ is correct. The resulting recall is consistently high, implying that most think-needed samples are successfully routed into the reasoning mode. Together, these findings demonstrate that the confidence of the initial answer provides a stable and reliable criterion for adaptive reasoning.

\begin{table}[t]
\begin{minipage}[t]{0.51\textwidth}
\centering
\footnotesize
\setlength{\tabcolsep}{0.33em}  
\caption{\textbf{Initial Answer's Confidence Separates Think-Needed Samples.} 
VideoMMMU shows markedly lower probability than MVBench and MMVU, indicating greater uncertainty. Accordingly, our confidence-based early exit triggers thinking more often on Video-MMMU, yielding a $+4\%$ accuracy gain.}    
\begin{tabular}{lccc}
\toprule
\textbf{Setting}    & \textbf{MVBench} & \textbf{MMVU} & \textbf{VideoMMMU}     \\
\midrule
\makecell[l]{Probability of Initial Answer}   & 0.948 & 0.933 & 0.874 \\
Think Ratio  &25\% & 39\% & 51\% \\
Performance Gains   & +0.1 & +0.4 & +4.0  \\ 
\midrule
\makecell[l]{Recall of Think-Needed Samples} & 100\% & 100\% & 94\% \\
\bottomrule
\end{tabular}
\label{table:dataset_prob}
\end{minipage}
\hfill
\begin{minipage}[t]{0.45\textwidth}
\centering
\footnotesize
\setlength{\tabcolsep}{0.3em}  
\caption{\textbf{Ablations on Reward Design.} Emphasizing the reviewed answer by setting $w_2>w_1$ outperforms equal weighting. Adding a small fallback reward $\alpha$ further improves accuracy.}
\vspace{-3pt}
\begin{tabular}{lcccccccc}
\toprule
\textbf{\makecell[l]{$w_1:w_2$}}   & \textbf{\makecell{$\alpha$}} & \textbf{VideoMME}    & \textbf{VideoMMMU} & \textbf{MVP} & \textbf{Charades-STA}  \\
\midrule
1:1 & \xmark & 66.1   & 56.1 & 38.3 & 58.3 \\
\midrule
0.9:1.1 & \xmark & 66.0   & 56.4 & 37.2 & 59.1 \\
\rowcolor{metabg}
0.9:1.1 & \cmark & \textbf{67.3}   & \textbf{58.6} & \textbf{39.4} & \textbf{60.0} \\
\midrule
0.8:1.2 & \xmark & 65.8     & 56.9 & 38.1 & 58.7 \\
0.8:1.2 & \cmark & 66.3   & 57.9 & 38.8 & 59.3 \\
\bottomrule
\end{tabular}
\label{table:ablation_reward}
\end{minipage}
\end{table}

\noindent\textbf{Ablation Study of Dual-Answer Reward Design.}
We ablate the dual-answer reward, a key component of our training framework, in Table~\ref{table:ablation_reward}. Since the model receives two verifiable rewards—one for the initial answer and one for the reviewed answer—their relative weighting is crucial. If $w_1\!=\!w_2$, the model may allow a correct $a_1$ to be overwritten by an incorrect $a_2$, so we assign $w_2\!>\!w_1$ to favor correctness in the final reviewed answer, especially when computation allows CoT reasoning. Empirically, asymmetric weights such as $0.9\!:\!1.1$ or $0.8\!:\!1.2$ outperform the uniform $1\!:\!1$ setting across multiple benchmarks.

We also study the fallback-tolerant reward, which discourages low-confidence guesses in $a_1$ and instead rewards honest deferral. As shown in the ablation, adding the fallback reward $\alpha$ consistently improves performance on reasoning benchmarks and achieves state-of-the-art results.

\noindent\textbf{Analysis of the Early-Exit Threshold.}
Figure~\ref{fig:inference_threshold} further studies the impact of the early-exit threshold $\tau$ on accuracy and think ratio under our confidence-based routing. As $\tau$ increases, early exit becomes more conservative, leading to a monotonic rise in the think ratio. Therefore, $\tau$ provides a direct and continuous control knob to trade efficiency for accuracy within a unified inference rule.

On reasoning-intensive benchmarks, higher $\tau$ consistently improves accuracy alongside increased reasoning usage. For VideoMMMU, rising $\tau$ from 0.86 to 0.98 improves accuracy from 57.5\% to 58.7\% while increasing the think ratio from 29\% to 55\%. Similarly on MVP, accuracy increases from 39.16\% to 39.37\% as the think ratio rises from 20\% to 51\%. These trends indicate that when the initial answer is less reliable, the reviewed-answer stage offers meaningful corrective benefits for these reasoning samples.

In contrast, on perception-oriented VideoMME, accuracy remains essentially unchanged across thresholds, whereas the think ratio still increases. This suggests diminishing returns from additional reasoning for easy perceptual queries. Based on these observations, we set $\tau=0.97$ as a robust default that preserves satisfied accuracy on reasoning-heavy tasks while limiting unnecessary CoT invocation on perception-heavy data, without requiring dataset-specific tuning.

\begin{figure*}[t]
  \centering
  \includegraphics[width=1.0\linewidth]{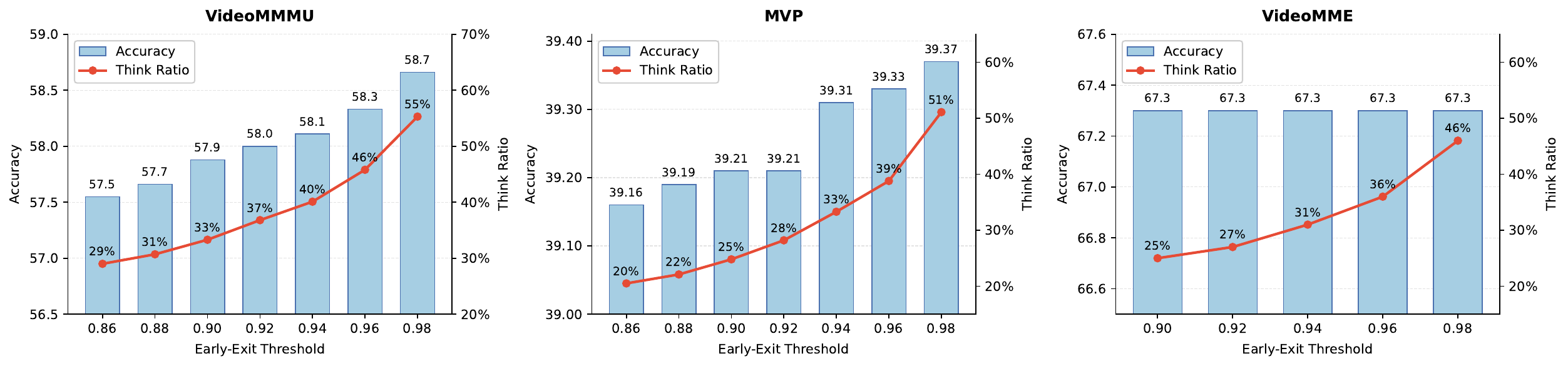}
    \caption{\textbf{Effect of the Early-Exit Threshold on Accuracy and Think Ratio.} In practice, we set $\tau = 0.97$ for all datasets.}
\label{fig:inference_threshold}
\end{figure*}

\noindent\textbf{Qualitative Result.} Figure~\ref{fig:examples_reasoning2} illustrates how \methodname{} leverages confidence-based early-exit to invoke reasoning only when needed. In this example, the model does not early-exit after the initial answer and instead performs advanced mathematical deduction, where it learns from the video to apply probability theory and integration. Although the initial prediction is D, the reviewed answer is revised to C after step-by-step reasoning, demonstrating the corrective value of the reasoning stage. More examples are provided in Appendix~\ref{appendix:examples}, further highlighting \methodname{}’s accuracy--efficiency balance.

\begin{figure*}[t]
    \centering
    \includegraphics[width=0.9\linewidth]{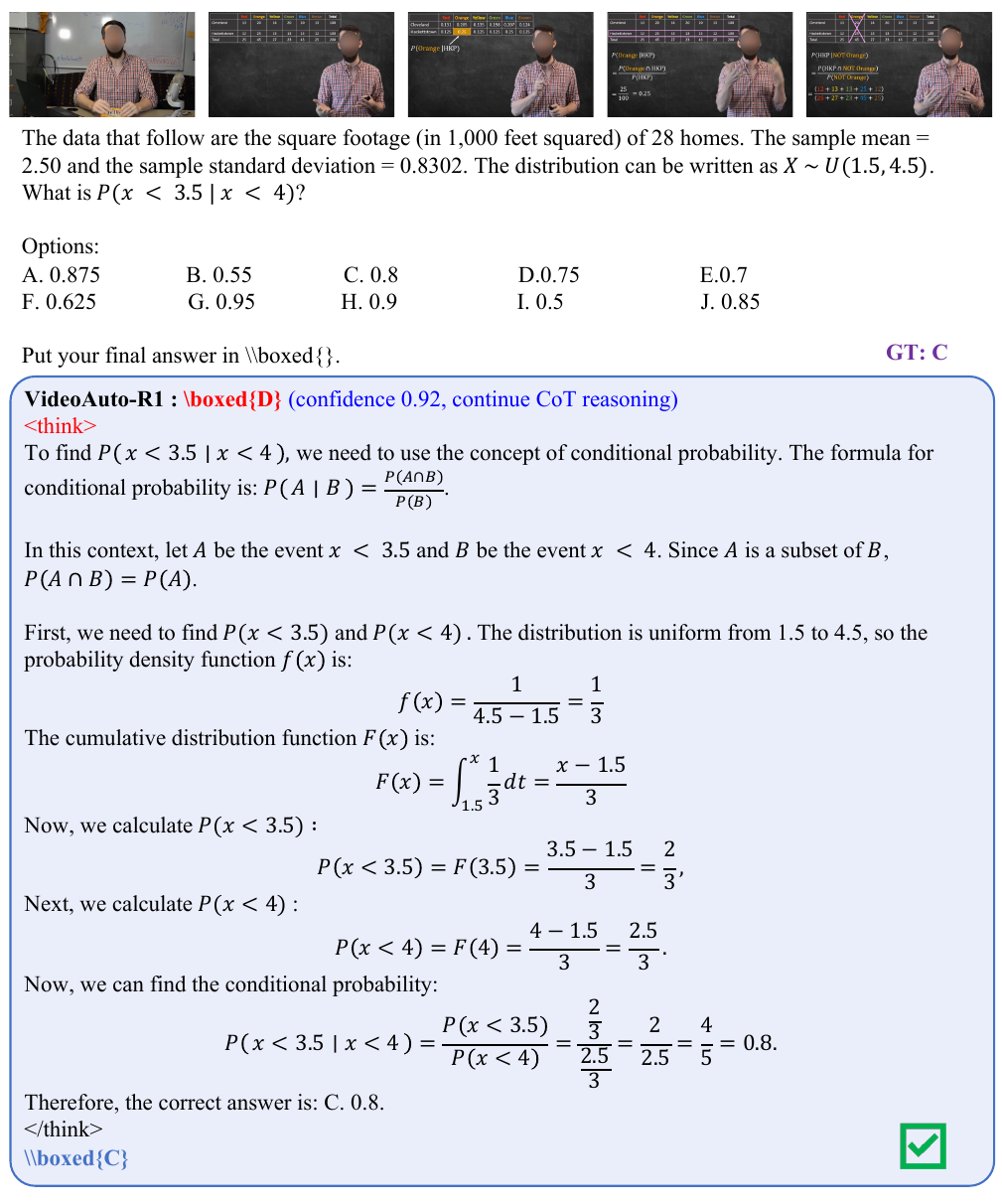}
    \caption{\textbf{\methodname{} Performing Complex Math Reasoning.} The model applies probability and integration, revising an incorrect initial answer to the correct one through structured reasoning.}
   \label{fig:examples_reasoning2}
\end{figure*}

\section{Conclusion}
We presented \methodname{}, an adaptive video reasoning framework that \emph{reasons only when necessary}. Motivated by the observation that long CoT does not reliably improve video understanding and can even degrade accuracy through overthinking, we proposed a thinking once, answering twice scheme to enable video auto-thinking.
Experiments on various video understanding benchmarks, such as perception, reasoning, and temporal grounding, consistently validate the advantages of our model. Our method is easy to formulate and implement, serving as an alternative to the standard reasoning framework. Our preliminary work suggests that \methodname{} has potential applications beyond video understanding.

\clearpage
\newpage
\bibliographystyle{assets/plainnat}
\bibliography{paper}

\clearpage
\newpage
\beginappendix

In this appendix, we provide more details of our method and present more experimental results. Specifically, we present the training data and its ablations in Section~\ref{appendix:data}. Next, we introduce the details of our reward design and related analysis in Section~\ref{appendix:reward_design}. After this, we present the prompt template used in our ablation experiments in Section~\ref{appendix:prompt_template}, as well as the training curve in Section~\ref{appendix:training_curve}. We also provide the algorithm details of our inference strategy in Section~\ref{appendix:inference_algorithm}. Then, we show additional experiments and further analysis in Section~\ref{appendix:additional_experiments}. Next, we discuss our limitations in Section~\ref{appendix:limitations}. Finally, we provide more examples for visualization and discussion in Section~\ref{appendix:examples}.

\section{Training Data}
\label{appendix:data}

\noindent\textbf{Data Composition.}
As described in the main paper, our training data consists of text, image, and video modalities.
For text-based reasoning, we incorporate DAPO-Math~\citep{yu2025dapo}; for image-based reasoning, we include ViRL~\citep{wang2025vl} and ThinkLite-Hard~\citep{wang2025sota}. For video QA, we draw from several sources including Video-R1~\citep{feng2025video}, TVBench~\citep{cores2024tvbench}, STI-Bench~\citep{li2025sti}, and MMR-VBench~\citep{zhu2025mmr}. To enhance temporal grounding and grounding-based QA capabilities, we additionally include Charades-STA~\citep{gao2017tall}, ActivityNet~\citep{caba2015activitynet}, Time-R1~\citep{wang2025time}, and NExT-GQA~\citep{xiao2024can}. 
All test samples from our evaluation benchmarks are manually excluded to prevent data leakage. The resulting training pool comprises approximately 137K samples.

\begin{table}[h]
    \centering
    \small
    \setlength{\tabcolsep}{2.5em}  
    \caption{\textbf{Training Dataset.} We include text, image, and video data during training, with a total of 83K samples.}
    \begin{tabular}{lll}
    \toprule
    \textbf{Type}   & \textbf{Size} & \textbf{Details} \\
    \midrule
    Text  & 6.4K & DAPO-Math~\citep{yu2025dapo} \\
    \midrule
    Image & 27.5K & ViRL~\citep{wang2025vl}, ThinkLite-Hard~\citep{wang2025sota} \\
    \midrule
    Video & 49.4K & \makecell[l]{Video-R1~\citep{feng2025video}, TVBench~\citep{cores2024tvbench}, \\ STI-Bench~\citep{li2025sti}, MMR-VBench~\citep{zhu2025mmr}, \\
    \midrule
    Charades-STA~\citep{gao2017tall}, ActivityNet~\citep{caba2015activitynet},  \\ Time-R1~\citep{wang2025time}, NExT-GQA~\citep{xiao2024can}} \\
    \bottomrule
    \end{tabular}
\label{table:data_details}
\end{table}

\begin{wrapfigure}{r}{0.45\textwidth}
\centering
\includegraphics[width=1.0\linewidth]{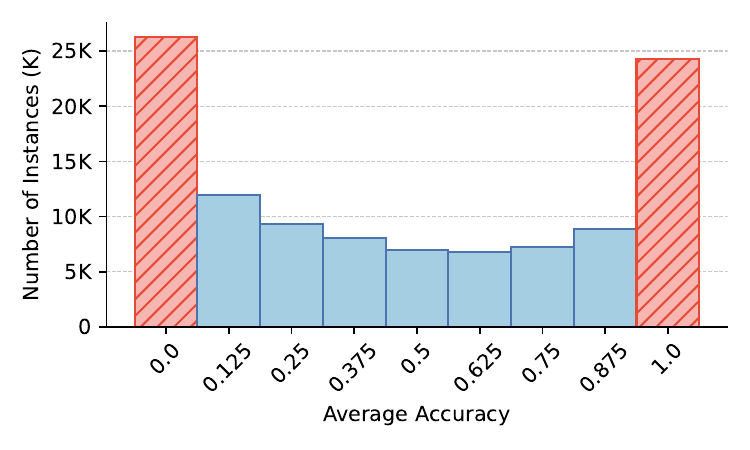}
\caption{\textbf{Distribution of per-sample accuracy} in the initial training pool, estimated by evaluating 8 diverse responses per sample. Samples with all responses correct or all incorrect are considered too easy or too hard and are excluded from QA-based data.}
\label{fig:data_distribution}
\end{wrapfigure}
\noindent\textbf{Filtering Pipeline.} 
We further curate a smaller, higher-quality subset from the initial data pool. First, we remove samples with invalid ground-truth (using \texttt{math-verify} for math problems and rule-based checks for QA problems). 
Next, for each remaining sample, we generate 8 responses using the base model (\ie, Qwen2.5-VL-7B-Instruct~\citep{bai2025qwen2}) with a high temperature. A smaller LLM (\ie, Qwen3-30B-A3B-Instruct~\citep{yang2025qwen3}) evaluates each response against the ground truth and assigns correct/incorrect labels. 
Samples for which all 8 responses are correct (too easy) or all are incorrect (too hard) are discarded, as they contribute little to GRPO-based reinforcement learning, as illustrated in Figure~\ref{fig:data_distribution}. 
This difficulty-based filtering is applied only to QA tasks; for temporal grounding, we retain all samples to mitigate the base model’s grounding weakness. After filtering, we finally obtain 83K samples. The detailed composition is listed in Table~\ref{table:data_details}.

\noindent\textbf{Effectiveness of Data Filtering.} 
To evaluate the effectiveness of our data filtering pipeline, we analyze the results presented in Table~\ref{table:data_performance}. Two key observations emerge from this analysis.
\textbf{First}, training solely on text data leads to a noticeable drop in performance on video tasks compared to the Qwen baseline, suggesting a domain shift and poor generalization. Adding image data significantly improves video QA performance, particularly on VideoMMMU, highlighting the importance of image-based math and reasoning data. However, due to the absence of temporal grounding data, performance on the Charades-STA benchmark remains low. When combining text, image, and video data, the model achieves the best overall performance under both filtered and unfiltered settings.
\textbf{Second}, in both the text+image and text+image+video configurations, removing overly easy or difficult samples leads to consistent performance gains. Additionally, this filtering reduces the number of training samples, thereby improving training efficiency. These findings validate the effectiveness of our data filtering strategy for GRPO-based reinforcement learning.

\begin{table*}[ht]
    \centering
    \small
    \caption{\textbf{Performance Comparison across Different Training Data and Filtering Strategy.} Note that we report the results under the RL with CoT setting. Combining text, image, and video data yields the best overall performance. Filtering out overly easy and hard samples consistently improves results while reducing dataset size, validating the effectiveness of our data curation pipeline.}
    \begin{tabular}{lccccccc}
    \toprule
    \textbf{Training Data}   & \textbf{Filtered} & \textbf{Size} & \textbf{VideoMME} & \textbf{MVBench} & \textbf{VideoMMMU}   & \textbf{Charades-STA}  \\
    \midrule
    Text & \xmark & 17K & 63.3 & 62.6 & 45.8 &  38.6 \\
    Image & \xmark & 50K & \textbf{65.6} & 66.8 & 52.8 & 40.1 \\
    Video & \xmark & 70K & 64.7 & \textbf{71.0} & \textbf{55.1} & \textbf{59.0} \\
    \midrule
    \multirow{2}{*}{Text + Image}  & \xmark & 67K & 66.1 & 67.4 & 53.3 & 41.6 \\
                         & \cmark & 34K & \textbf{67.0} & \textbf{68.5} & \textbf{56.4} & \textbf{42.0} \\
    \midrule
    \multirow{2}{*}{Text + Image + Video} & \xmark & 138K & 65.4 & 71.0 & 55.4 & 59.7 \\
     & \cmark & 83K & \textbf{66.1} & \textbf{71.2} & \textbf{56.4} & \textbf{59.8} \\
    \bottomrule
    \end{tabular}
\label{table:data_performance}
\end{table*}

\section{Reward Designs}
\label{appendix:reward_design}

To complement the reward description in the main paper, we provide the details below. Our overall reward is defined as a weighted sum of the task reward and the format reward.

\noindent\textbf{Task Reward.}
We consider three task types for computing task rewards: QA, temporal grounding, and grounding QA.

\begin{itemize}

\item \textit{Question Answering.}
For math problems, we use \texttt{math-verify} to compare the prediction with the ground truth; otherwise we compare normalized strings (\eg, case-folded, whitespace stripped). This yields a binary reward
\[
R_{\text{QA}}(o_i) \in \{0,1\}.
\]
\item \textit{Temporal Grounding.}
Let the ground-truth segments be $\mathcal{G}=\{[s_j,e_j]\}_j$ and the predicted segments be $\widehat{\mathcal{G}}=\{[\hat{s}_k,\hat{e}_k]\}_k$ (either set may contain one or multiple segments). We compute the temporal IoU and take the best matching pair with the largest tIoU. If no valid segment can be parsed, we assign $R_{\text{TG}}(o_i) = 0$.
\[
R_{\text{TG}}(o_i) \;=\; 
\max_{[\hat{s},\hat{e}]\in \widehat{\mathcal{G}},\;[s,e]\in \mathcal{G}}
\mathrm{tIoU}\!\left([\hat{s},\hat{e}],[s,e]\right)\;\in [0,1],
\]
\item \textit{Grounding QA.}
We parse the textual answer and the predicted segments from the model output, compute $R_{\text{QA}}(o_i)$ and $R_{\text{TG}}(o_i)$ as above, and sum them:
\[
R_{\text{GQA}}(o_i) \;=\; R_{\text{QA}}(o_i) + R_{\text{TG}}(o_i)\;\in [0,2].
\]
\end{itemize}

\noindent\textbf{Format Reward.} 
In addition to task correctness, we use a binary format reward $R_{\text{fmt}}(o_i)\in\{0,1\}$ enforced via strict regex checks. 
For \methodname{},  we require exactly two \verb|\\boxed{...}| answers, and in between one \texttt{<think>...</think>} block, with no additional text before, between, or after.

\noindent\textbf{Analysis of the Dual-Answer Reward Design.}
\label{appendix:reward_analysis}
In Section~4.2 of the main paper, we introduce the dual-answer reward design used during training. The key components of this design are the weight coefficients $w_1$ and $w_2$ assigned to the initial and reviewed answers, respectively, as well as the fallback bonus weight $\alpha$. Table~\ref{table:reward_list} summarizes the effects of different choices for these coefficients.

\begin{wraptable}{r}{0.45\textwidth}
\vspace{-1em}
\centering
\small
\setlength{\tabcolsep}{0.28em}  
\caption{\textbf{Effects of Dual-Answer Reward Coefficients.}}
\begin{tabular}{ccccc}
\toprule
\textbf{\makecell{First \\ Answer}}  & \textbf{\makecell{Second \\Answer}}  & \makecell{$w_1\!=\!1$,\\$w_2\!=\!1$,\\$\alpha\!=\!0$} & \makecell{$w_1\!=\!0.9$,\\$w_2\!=\!1.1$,\\$\alpha\!=\!0$} & \makecell{$w_1\!=\!0.9$,\\$w_2\!=\!1.1$,\\$\alpha\!=\!0.3$}\\
\midrule
\xmark & \xmark                    & 0 & 0 & 0 \\
\textit{Let's analyze...} & \xmark & 0 & 0 & 0 \\
\rowcolor{metabg}
\cmark & \xmark                    & 1 & 0.9 & 0.9 \\
\rowcolor{metabg}
\xmark & \cmark                    & 1 & 1.1 & 1.1 \\
\rowcolor{metabg}
\textit{Let's analyze...} & \cmark & 1 & 1.1 & 1.4 \\
\cmark & \cmark                    & 2 & 2 & 2\\
\bottomrule
\end{tabular}
\label{table:reward_list}
\vspace{-1em}
\end{wraptable}
First, when $w_1 = w_2$, the model assigns identical rewards to two distinct cases: (i) the first answer is correct but the second is wrong, and (ii) the first answer is wrong but the second is correct. However, our intention is to prioritize the correctness of the \emph{reviewed answer}, since users who permit step-by-step reasoning with a sufficient compute budget expect the final answer to be reliable. Therefore, equal weighting fails to distinguish these two scenarios. By choosing $w_1\!<\!w_2$ (\eg, $0.9\!:\!1.1$), the total reward becomes $0.9$ for a ``correct $\rightarrow$ wrong’’ pattern, but $1.1$ for ``wrong $\rightarrow$ correct’’, thereby encouraging the model to produce accurate reviewed answers during RL.

Second, even with $w_1 < w_2$, the model still assigns the same reward when the first output is an incorrect guess or a fallback string \textit{``Let's analyze the problem step-by-step.''} The fallback string is not a wrong prediction; rather, it is an explicit and honest signal that the model identifies the task as difficult and intentionally defers reasoning to the next stage. Such behavior should be incentivized. By introducing the fallback bonus $\alpha$, as shown in the last column of Table~\ref{table:reward_list}, the model is able to clearly differentiate between an incorrect guess and a fallback indicator.

Finally, when both the initial and reviewed answers are correct, the model receives the highest possible reward, which aligns with our design goal.

\section{Prompt Template}
\label{appendix:prompt_template}

In the main paper, we introduce the system prompt used in \methodname{}, which adopts an \emph{answer $\rightarrow$ think $\rightarrow$ answer} format. This prompt design avoids a cold-start stage and facilitates stable training with promising performance. Additionally, in Table 5 of the main paper, we explore alternative reinforcement learning settings.

\noindent\textbf{RL without Thinking.}
As shown in Table~\ref{table:rl_direct_prompt}, this variant directly applies GRPO without requiring any intermediate explanation. The model is prompted to provide only the final answer enclosed in a \verb|\\boxed{}| command.

\noindent\textbf{RL with Thinking.}
As shown in Table~\ref{table:rl_think_prompt}, this is the standard prompt for GRPO training. The model first generates a reasoning trace within \verb|<think> </think>| tags, followed by the final answer enclosed in \verb|\\boxed{}|. This prompt format aligns with previous R1-style approaches such as Video-R1~\citep{feng2025video} and VideoChat-R1~\citep{li2025videochat}.

\begin{table}[ht]
    \centering
    \footnotesize
    \caption{\textbf{System Prompt for RL without Thinking.}}    
        \begin{tabular}{p{16cm}}
        \toprule
        \textbf{\textit{SYSTEM PROMPT}} \\
        \midrule
        You are a helpful assistant. Put your final answer in {\color{blue}\verb|\\boxed{}|}. \\
        \bottomrule
    \end{tabular}
\label{table:rl_direct_prompt}
\end{table}

\begin{table}[ht]
    \centering
    \footnotesize
    \caption{\textbf{System Prompt for RL with Thinking.}}    
        \begin{tabular}{p{16cm}}
        \toprule
        \textbf{\textit{SYSTEM PROMPT}} \\
        \midrule
        You are a helpful assistant. \\
        FIRST, think through the reasoning process as an internal monologue, and THEN provide the final answer. The reasoning process MUST be enclosed within {\color{RedOrange}\verb|<think> </think>|} tags, and the final answer MUST be wrapped in {\color{blue}\verb|\\boxed{}|}. \\
        \bottomrule
    \end{tabular}
\label{table:rl_think_prompt}
\end{table}

\section{Training Curve}
\label{appendix:training_curve}

To better understand the behavior of \methodname{}, we visualize the training curves of the task rewards for both the initial and reviewed answers during training, as shown in Figure~\ref{fig:training_curve}. We highlight three key observations below.

\noindent\textbf{Reviewed Answer \textit{vs}. Initial Answer.}
For both Qwen2.5-VL-7B and Qwen3-VL-8B, the reviewed answer consistently achieves a higher task reward than the initial answer during training. This performance gap remains stable after convergence, indicating that the \emph{answer–think–answer} paradigm effectively leverages intermediate reasoning to refine predictions. Moreover, this confirms that the dual-answer reward design (with $w_2 \!>\! w_1$) can encourage the model to treat the second answer as a meaningful revision rather than a naive re-sampling of the first.

\begin{figure}[t]
\centering
\includegraphics[width=0.9\linewidth]{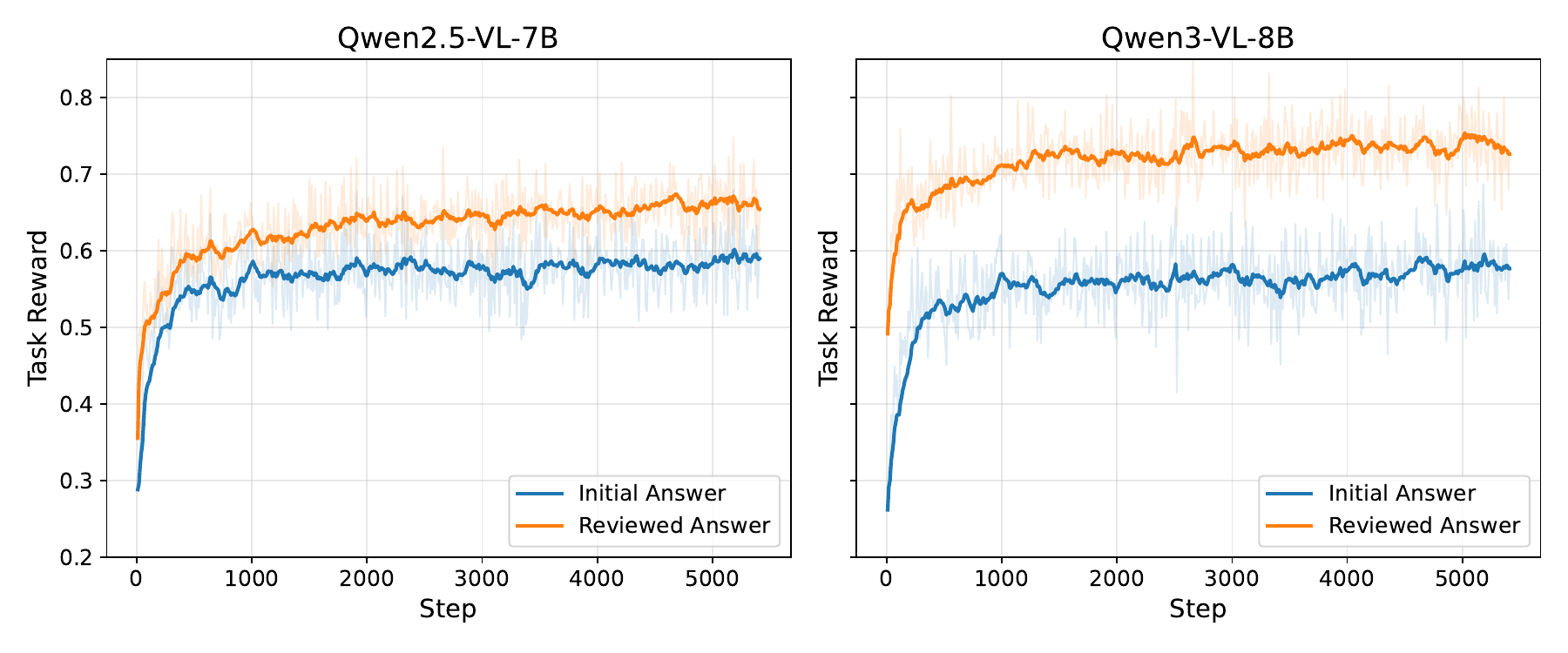}
\vspace{-12pt}
\caption{\textbf{Training Curves of \methodname{}.} We show the average task reward for both initial and reviewed answers during GRPO training.}
\label{fig:training_curve}
\vspace{-8pt}
\end{figure}

\noindent\textbf{Training Dynamics.}
As training progresses, the task rewards for both answers increase. In the early stages, we observe a rapid improvement, followed by a slower but steady rise until convergence. This pattern suggests that GRPO quickly captures coarse task structure and gradually optimizes finer-grained reasoning capabilities over time.

\noindent\textbf{Impact of Backbone Capacity.}
Throughout training, Qwen3-VL-8B consistently outperforms Qwen2.5-VL-7B in both answers. The stronger backbone benefits from better initialization and sustains a higher reward margin after convergence. These results demonstrate that \methodname{} scales effectively with model capacity: larger base models can more fully exploit dual-answer supervision and confidence-based reasoning, resulting in higher final results.

\section{Inference Strategy}
\label{appendix:inference_algorithm}

\begin{algorithm}[ht]
\caption{Inference Strategy of \methodname{}}
\label{alg:inference}
\begin{algorithmic}[1]
\Require Trained model $p_\theta$, video $v$, question $q$, confidence threshold $\tau$, fallback string $f$
\Ensure Predicted answer $\hat{a}$
\State Given input $(v, q)$, perform greedy decoding until the first \verb|<think>| tag is generated.
\State Let $a_1 = (t_1, \dots, t_L)$ be the tokens inside the first box, and let $y_{\leq \ell_0}$ denote the prefix up to (and including) the opening of $a_1$
\Statex 
\If{$a_1 = f$}  \Comment{{\color{blue}designated fallback string}}
    \State $s(a_1) \gets -1e6$
\Else
    \State Compute length-normalized confidence
    $s(a_1) \gets \frac{1}{L} \sum_{\ell=1}^{L} 
        \log p_\theta\!\big(t_\ell \mid y_{\leq \ell_0+\ell-1}, x\big)$
\EndIf
\Statex 
\If{$s(a_1) \ge \log \tau$} \Comment{{\color{blue}early exit}}
    \State Accept the initial answer
    \State \Return $\hat{a} \gets a_1$
\Else \Comment{{\color{blue}continue reasoning}}
    \State Resume decoding from the current prefix
    \State Generate rationale $r$ enclosed in \verb|<think>...| \verb|</think>| and the second boxed answer $a_2$
    \State \Return $\hat{a} \gets a_2$
\EndIf
\end{algorithmic}
\end{algorithm}

At test time, \methodname{} employs a confidence-based early-exit mechanism to determine whether to stop after generating the initial direct answer or to proceed with a full chain-of-thought rationale followed by a reviewed answer. Algorithm~\ref{alg:inference} summarizes this procedure, which consists of three main steps: (1) generate the initial answer, (2) compute its confidence score, and (3) decide whether to exit early or continue reasoning.

For implementation simplicity, we terminate generation early by detecting the appearance of the opening \verb|<think>| tag during greedy decoding. We then extract the token sequence enclosed in the first \verb|\\boxed{}| block, which always precedes the \verb|<think>| tag. Since the initial answer $a_1$ typically consists of only a few tokens, this strategy enables low-overhead confidence computation while providing substantial savings in decoding latency and token budget whenever early exit is triggered.

\begin{table*}[t]
    \centering
    \small
    \setlength{\tabcolsep}{0.24em}  
    \caption{\textbf{Evaluation Results on Video QA Benchmarks with Different Frames.} For the Qwen2.5-VL models, we allow up to 16K total video tokens. For the Qwen3-VL models, we allow up to 128K total video tokens.}
    \begin{tabular}{lcccccccc}
    \toprule
    \multirow{3}{*}{\textbf{Model}}                     & \multirow{3}{*}{\textbf{Frames}}    & \multicolumn{4}{c}{\textbf{Video Perception Benchmark}} & & \multicolumn{2}{c}{\textbf{Video Reasoning Benchmark}} \\
    \cmidrule{3-6}\cmidrule{8-9}
                            &    & VideoMME             & MVBench             & LongVideoBench    & MMVU     &     & VideoMMMU             & MVP  \\
    \midrule
    Qwen2.5-VL-7B & 64 & 63.1 & 67.0 & 59.7 & \textbf{66.2}  & & \textbf{54.6} & 35.8\\
    Qwen2.5-VL-7B & 128 & 65.9 & 67.0 & 60.6 & 66.2 &  & 54.7 & 35.8\\
    Qwen2.5-VL-7B & 256 & \textbf{66.0} & \textbf{67.1} & \textbf{60.9} & 65.7 & & 52.7  & \textbf{36.5} \\
    \midrule
    \textbf{\methodname{}}\tiny{(Qwen2.5-VL-7B)}  & 64 &  64.6 & 71.0 & 60.0 & \textbf{69.7} & & \textbf{58.7} & 39.2 \\
    \textbf{\methodname{}}\tiny{(Qwen2.5-VL-7B)}  & 128 &  66.7 & 71.0 & 60.4 & 69.1 & & 56.6 & 39.3 \\
    \textbf{\methodname{}}\tiny{(Qwen2.5-VL-7B)}  & 256 & \textbf{67.3} & \textbf{71.0} & \textbf{60.5} & 68.6 & & 56.7  & \textbf{39.4} \\
    \midrule
    Qwen3-VL-8B & 64 &  67.3 & 69.4 & 63.4 & 69.9 & & \textbf{61.0} & 40.4 \\
    Qwen3-VL-8B & 256 & 70.9 & 69.4 & 66.0 & 69.6 & & 59.9 & 40.5  \\
    Qwen3-VL-8B & 2048 & \textbf{72.5} & \textbf{69.4} & \textbf{67.6} & \textbf{69.9} & & 59.8  & \textbf{40.5} \\
    \midrule
    \textbf{\methodname{}}\tiny{(Qwen3-VL-8B)} & 64  & 67.9 & 71.8 & 63.9 & 71.0 & & \textbf{65.0} & 42.7 \\
    \textbf{\methodname{}}\tiny{(Qwen3-VL-8B)} & 256  & 70.4 & 72.0 & 67.1 & 71.0 & & 63.8 & 42.9\\
    \textbf{\methodname{}}\tiny{(Qwen3-VL-8B)} & 2048 & \textbf{71.7} & \textbf{72.0} & \textbf{67.4} & \textbf{71.1} & & 64.0 & \textbf{43.0} \\
    \bottomrule
    \end{tabular}
\label{table:benchmark_qa_frames}
\end{table*}

\section{Additional Experiments}
\label{appendix:additional_experiments}

In this section, we present additional experiments and analyses to complement the findings reported in the main paper.

\subsection{Performance with Different Frames}
\label{appendix:performance_of_different_frames}

In the main paper, we report the best-performing configurations of our model. Here, we present the complete results in Table~\ref{table:benchmark_qa_frames} and analyze how the number of input frames affects performance on both perception and reasoning benchmarks.

Under a 16K video-token budget using Qwen2.5-VL, increasing the number of frames from 64 to 256 yields noticeable improvements on most perception benchmarks for both the Qwen baseline and \methodname{}. For example, accuracy on VideoMME improves from 63.1\% to 66.0\%, and on LongVideoBench from 59.7\% to 60.9\%. However, the reasoning-oriented benchmark VideoMMMU shows weaker dependence on frame count, where performance slightly decreases with additional frames. This trend persists when switching to Qwen3-VL, which supports a larger 128K video-token budget and up to 2{,}048 frames.

Moreover, \methodname{} achieves consistent improvements compared to the Qwen baseline. For instance, even under a 64-frame budget, \methodname{} improves upon the baseline performance from 63.1\% to 64.6\% on VideoMME, and from 66.2\% to 69.7\% on MMVU, demonstrating the effectiveness of our proposed approach across both low and high frame regimes.

\begin{table*}[t]
    \centering
    \small
    \setlength{\tabcolsep}{1.2em}  
    \caption{\textbf{Comparison of Different Inference Strategies on Temporal Grounding Benchmarks.} We compare the results using the first boxed answer, the second boxed answer, or the confidence-based early-exit answer. We observe that on grounding benchmark, the first boxed answer is typically sufficient, so we early-exit without further reasoning to save computation.}
    \begin{tabular}{lccccccccc}
    \toprule
    \multirow{2}{*}{\textbf{Model}} &  \multirow{2}{*}{\textbf{\makecell{Inference Strategy}}}          & & \multicolumn{4}{c}{\textbf{ActivityNet}} & & \multicolumn{2}{c}{\textbf{NExT-GQA}} \\
     & &&  0.3 & 0.5 & 0.7 & mIoU & & Acc & mIoU\\
    \midrule
    \multirow{3}{*}{\makecell{\textbf{\methodname{}}\\\tiny{(Qwen2.5-VL-7B)}}} & First Answer & & 69.2 & 48.5 & 27.3 & 47.6 & & 80.6 & 36.7 \\
     & Second Answer & & 69.2 & 48.5 & 27.3 & 47.6 & & 80.6 & 36.7 \\
     & Auto & & 69.2 & 48.5 & 27.3 & 47.6 & & 80.6 & 36.7 \\
    \bottomrule
    \end{tabular}
\label{table:benchmark_grounding_detail}
\vspace{8pt}
\end{table*}

\subsection{Analysis on Temporal Grounding Benchmarks}
\label{appendix:analysis_on_grounding}

In the main paper, we emphasize that for grounding benchmarks, the initial answer is typically sufficient, so we exit early by default to save computation. In Table~\ref{table:benchmark_grounding_detail}, we report the detailed grounding results when using the first boxed answer, the second boxed answer, and the confidence-based auto strategy.

\noindent\textbf{Initial \textit{vs.}\ Reviewed Answer.}
Unlike video QA benchmarks, temporal grounding shows almost no gap between the first and reviewed answers. For \methodname{}, mIoU is the same for ActivityNet and NExT-GQA when comparing the first and second boxed answers. On NExT-GQA, the grounding QA accuracy also remains the same.

We hypothesize two reasons for this phenomenon. \textbf{First}, since the grounding procedure does not require multi-step logical deduction, the model can map the queried event to a time span directly from perception. Once the model has localized a segment in the first answer, additional textual reasoning has limited room to further improve the IoU. \textbf{Second}, since we lack the SFT stage to teach the model how to explicitly reason on the grounding task, the model cannot easily refine the predicted segments. Consequently, the reasoning stage rarely corrects localization errors, leading to nearly identical scores. In practice, this suggests that for grounding tasks, RL still shows significant improvements compared to baseline or SFT, but it is often unnecessary to rely on long and language-based thinking rationales.

\noindent\textbf{Reasoning Traces on QA vs.\ Grounding.}
To better understand this behavior, we examine representative reasoning traces of \methodname{} between grounding and QA tasks, as shown in Figure~\ref{fig:examples_grounding},~\ref{fig:examples_qa}, and ~\ref{fig:examples_reasoning2}. On video QA benchmarks, the thinking rationale usually contains multi-step analysis: enumerating visual evidence, performing arithmetic, or checking answer options. In contrast, grounding traces are much shorter. The model typically identifies the relevant event or shot, notes when it appears and disappears in the video, and then outputs the corresponding timestamps or intervals.

These qualitative observations align with the quantitative results in Table~\ref{table:benchmark_grounding}: for temporal grounding benchmarks, explicit reasoning provides limited additional benefit over the direct localization. Therefore,  we use the direct answering results on grounding benchmarks for \methodname{}.

\subsection{Analysis of the Impact of Cold-Start SFT}
\label{cold_start_ablation}

In our training framework, we deliberately omit chain-of-thought SFT and proceed directly to RL. Traditionally, SFT is used to (1) teach the CoT output format, (2) imitate the CoT reasoning process, and (3) acquire general knowledge from newly collected data. However, with modern base models that are already trained on massive corpora, the marginal benefit for (1) and (3) is limited. Moreover, collecting large-scale, high-quality CoT traces for (2) is expensive and often noisy.

\begin{table}[hb]
    \centering
    \small
    \caption{\textbf{Ablation on Cold-Start CoT SFT.}}
    \begin{tabular}{lccc}
    \toprule
    \textbf{Setting}  & \textbf{VideoMME}  & \textbf{MVBench} & \textbf{VideoMMMU} \\
    \midrule
    Qwen2.5-VL baseline &   66.0  & 67.1 & 54.7 \\
    SFT with Video-R1-CoT data & 60.1 & 64.0 & 53.8 \\
    \midrule
    RL with thinking & 66.1 & 71.2 &  56.4 \\
    SFT $\rightarrow$ RL with thinking & 61.7 & 64.3 & 53.5 \\
    \bottomrule
    \end{tabular}
\label{table:cold_start_ablation}
\end{table}

In early experiments, SFT on Video-R1-CoT data~\citep{feng2025video}, which has both the intermediate reasoning traces and final answer, not only failed to improve performance, but actually degraded the Qwen2.5-VL baseline, a phenomenon also observed in prior work~\citep{li2025veripo,chen2025sft}. Table~\ref{table:cold_start_ablation} summarizes this effect. Pure SFT substantially hurts performance across all three benchmarks. When we apply GRPO starting from the SFT checkpoint (``SFT $\rightarrow$ RL with thinking''), the final model remains significantly worse than RL applied directly on the base model.

These results suggest that low-quality CoT supervision can distort the behavior of a strong base model and create a poor initialization for RL. We therefore focus on directly incentivizing the base model’s reasoning via GRPO-style reinforcement learning.

\section{Limitations}
\label{appendix:limitations}

In this section, we mainly discuss three limitations of our work and leave them as future work.

\textbf{First}, our distinction between direct answering and reasoning is currently made purely at test time via a confidence-based early-exit rule on the first boxed answer. While this mechanism is simple and effective, it does not explicitly shape the confidence distribution during training. A natural extension would be to incorporate the probability of the first boxed answer into the training objective itself: for simple questions, the model should be encouraged to assign high confidence to a correct direct answer, whereas for genuinely hard questions it should learn to keep the initial confidence low and defer to the reasoning stage. Jointly optimizing both accuracy and calibrated confidence could further improve the reliability of the early-exit policy. 

\textbf{Second}, our current reasoning mechanism relies strictly on language-based chain-of-thought. While effective for symbolic and logical tasks, we observe that such textual reasoning yields limited improvements on perception-oriented QA and temporal grounding benchmarks compared to direct answering. This suggests that purely semantic rationales may be insufficient to correct fine-grained visual perception errors or refine precise temporal boundaries once the initial visual encoding is fixed. Future work could explore interleaved multimodal reasoning paradigms, such as "thinking with frames", where the model explicitly revisits video segments or visual features during the reasoning to enhance perceptual precision and grounding accuracy.

\textbf{Third}, the existing video reasoning benchmarks are still limited in scope and difficulty. Many datasets contain relatively short clips and perception-oriented questions. More advanced benchmarks that stress long-range temporal dependencies, compositional logic, and counterfactual reasoning, rather than just math or symbolic-heavy problems, are needed to more faithfully evaluate and compare the reasoning capabilities of MLLMs.

\begin{figure*}[t]
    \centering
    \includegraphics[width=0.95\linewidth]{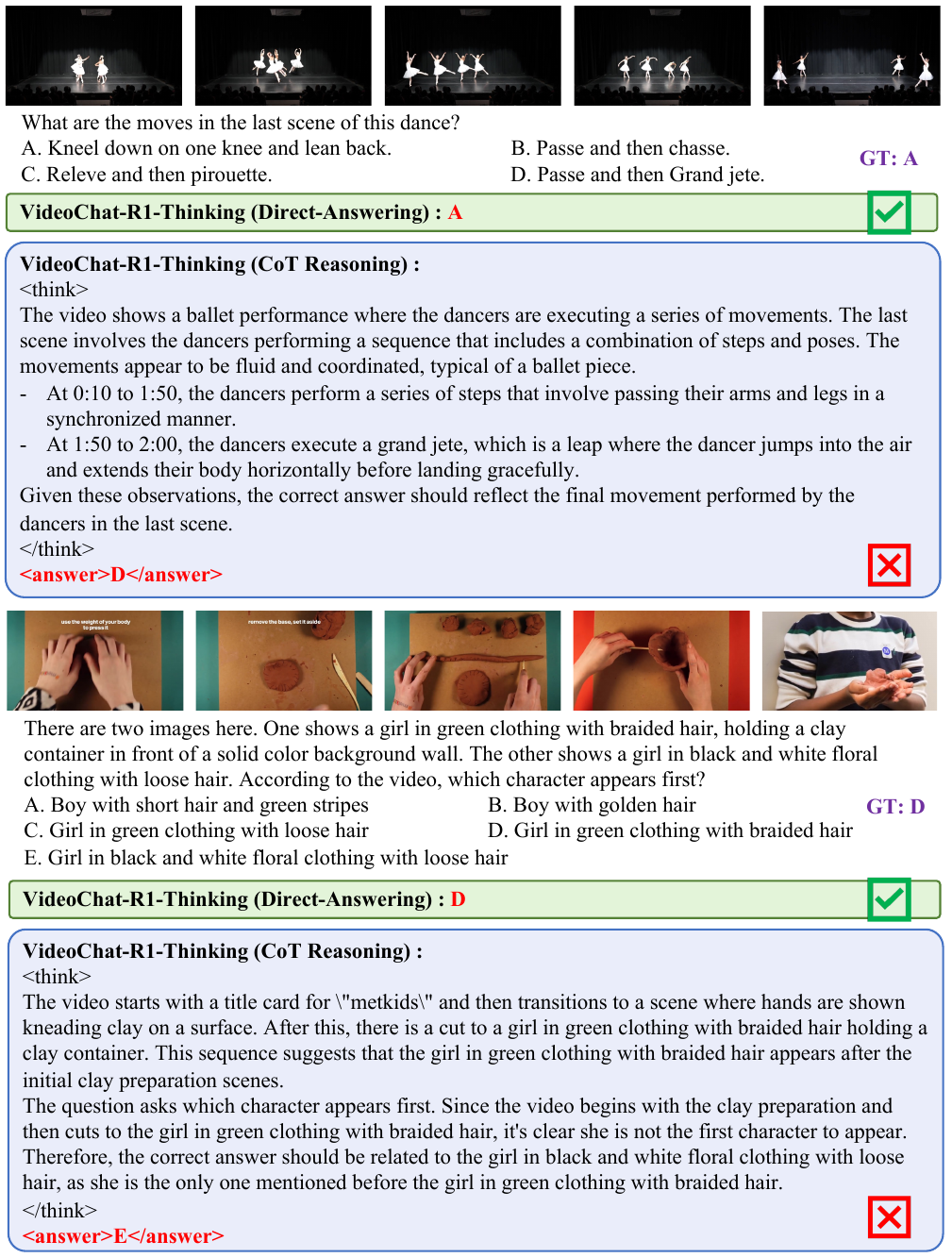}
    \caption{\textbf{Failure case of VideoChat-R1.} The model hallucinates visual content in its reasoning trace, leading to an incorrect answer. Direct answering performs better in this perception-oriented task.}
   \label{fig:examples_videochat_fail}
\end{figure*}

\begin{figure*}[t]
    \centering
    \includegraphics[width=0.95\linewidth]{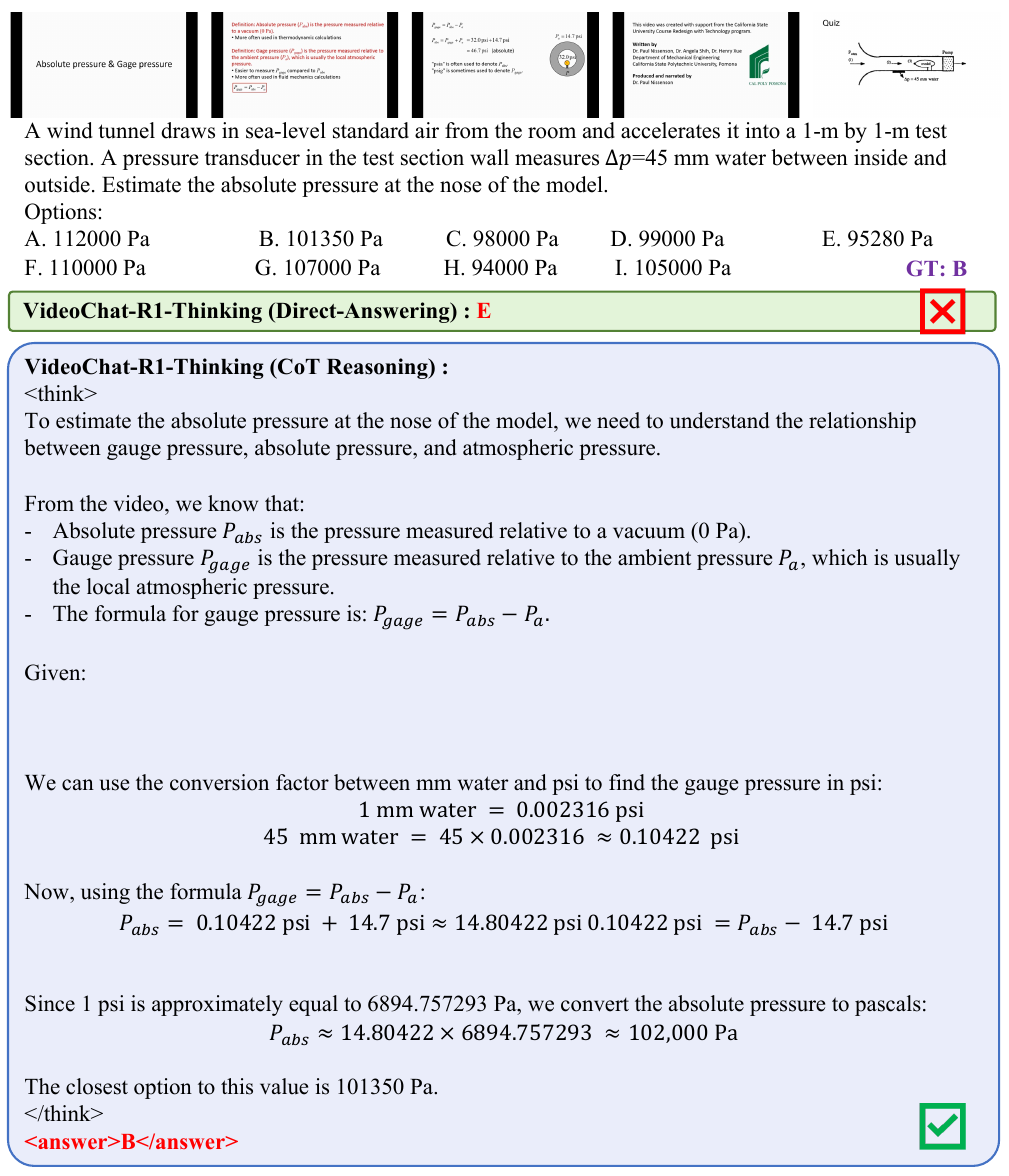}
    \caption{\textbf{Success Case of VideoChat-R1 on VideoMMMU.} The question involves scientific analysis with equations. Chain-of-thought reasoning leads to the correct answer, whereas direct answering would likely fail.}
   \label{fig:examples_videochat_fail_success}
\end{figure*}

\begin{figure*}[t]
    \centering
    \includegraphics[width=0.97\linewidth]{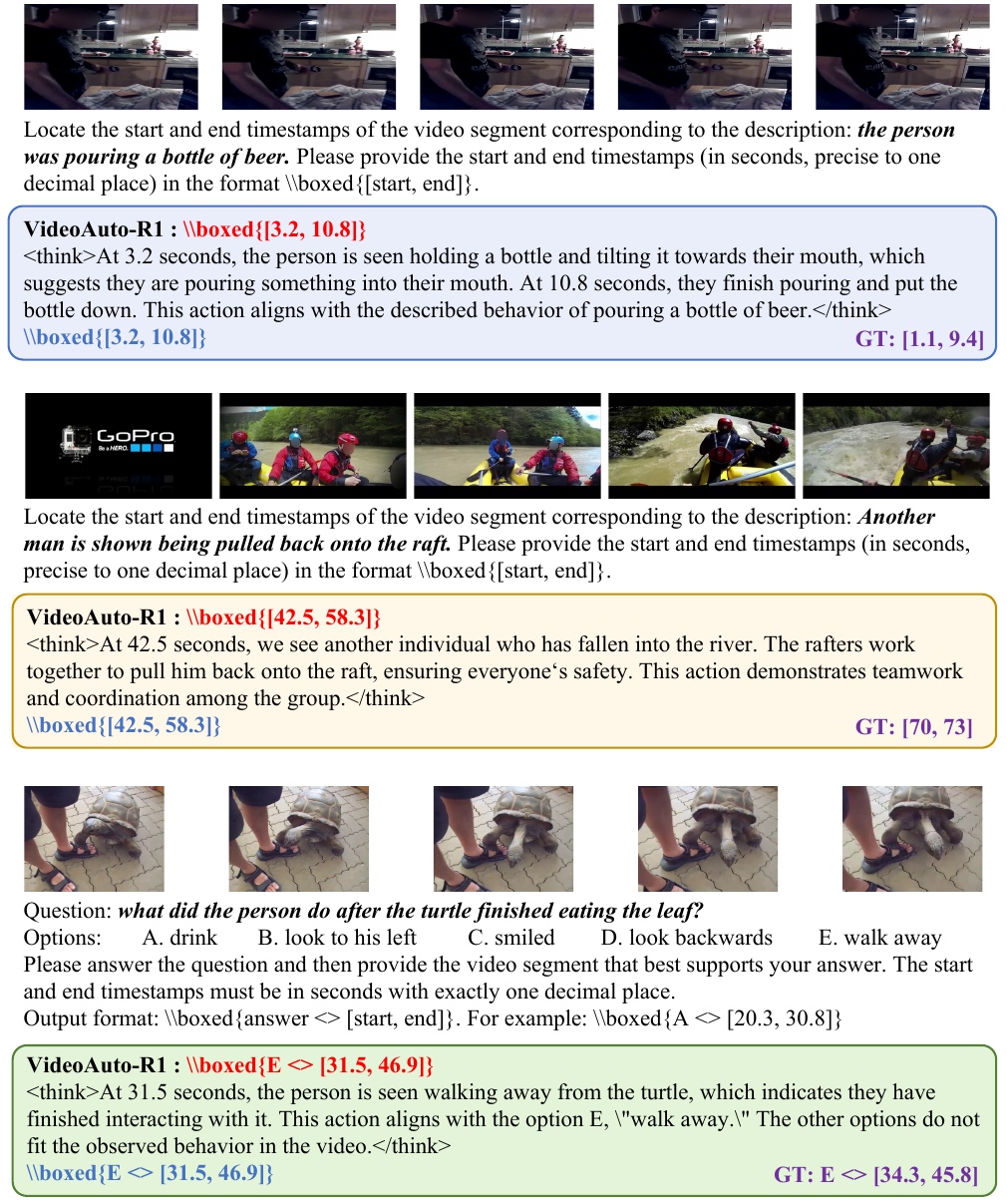}
    \caption{\textbf{\methodname{} on Temporal Grounding Tasks.} The reasoning trace is simple and redundant with the initial answer, enabling effective early-exit without full CoT reasoning.}
   \label{fig:examples_grounding}
\end{figure*}

\begin{figure*}[t]
    \centering
    \includegraphics[width=0.97\linewidth]{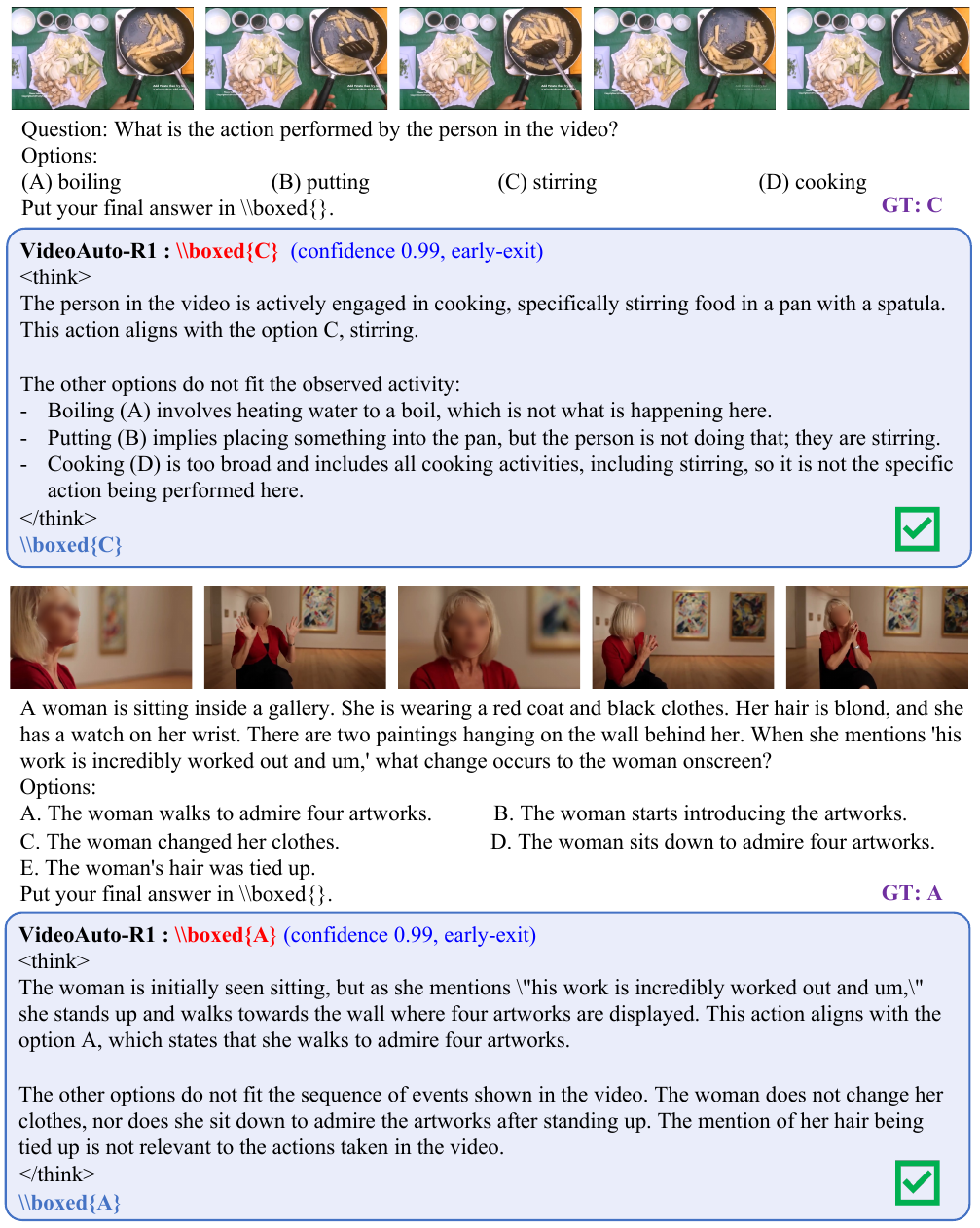}
    \caption{\textbf{\methodname{} on Perception-Oriented QA Tasks.} High-confidence initial answers trigger early exit, improving inference efficiency.}
   \label{fig:examples_qa}
\end{figure*}

\begin{figure*}[t]
    \centering
    \includegraphics[width=0.97\linewidth]{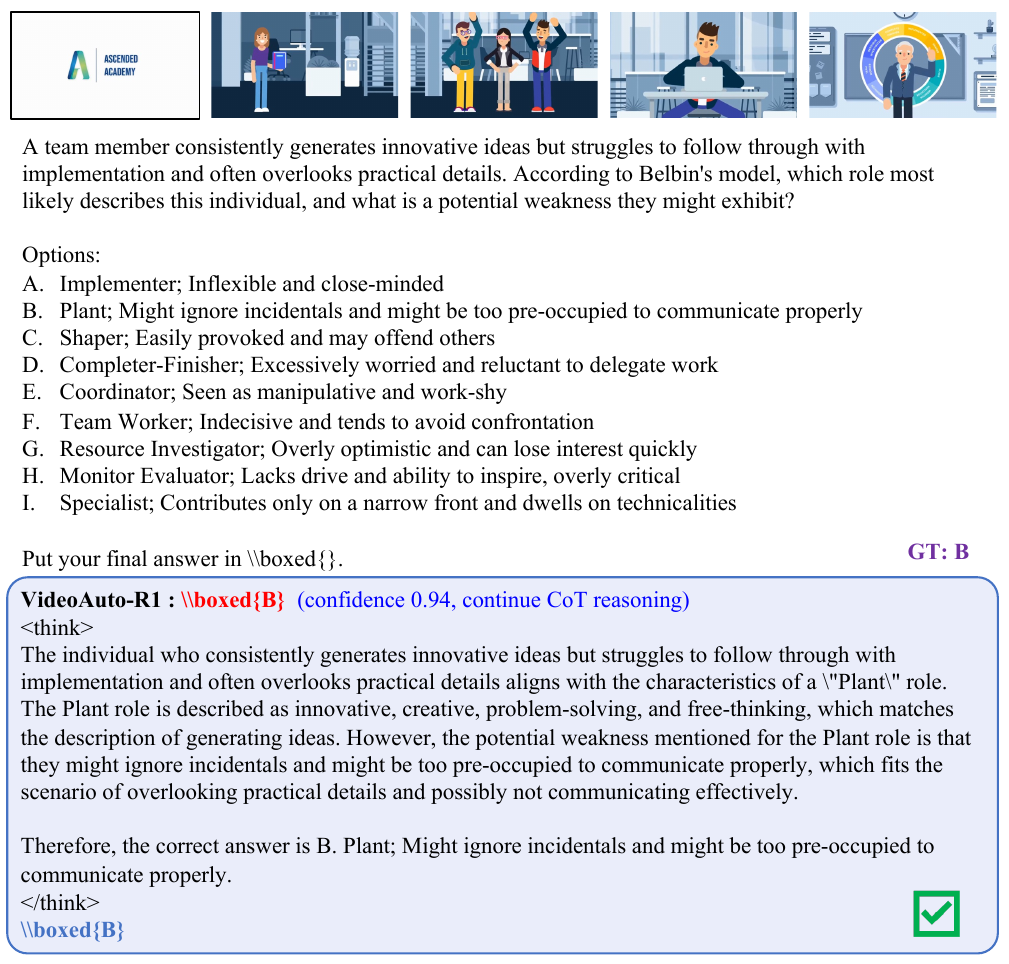}
    \caption{\textbf{\methodname{} on Reasoning-Oriented QA Tasks.} The reasoning trace is longer and more detailed, with clear step-by-step deductions.}
   \label{fig:examples_reasoning1}
\end{figure*}

\textbf{Fourth}, truly ``\textit{must-think}'' video data, where multi-step reasoning is indispensable rather than merely helpful, remains scarce. Constructing high-quality, large-scale video datasets that explicitly require deep reasoning (for example, multi-event causal chains, non-trivial temporal puzzles, or physically challenging scenarios) is therefore an urgent and valuable direction for future work. In the meantime, exploring the advanced reasoning pattern for the grounding task is also an interesting direction.

\section{Qualitative Examples}
\label{appendix:examples}

In this section, we provide additional qualitative results to support our analysis.

In Figure~\ref{fig:examples_videochat_fail}, we first present a failure case of VideoChat-R1~\citep{li2025videochat}, where the direct answer is correct but the CoT-reasoned result is incorrect. Although the model generates a seemingly reasonable step-by-step rationale, it suffers from hallucinations. For example, it mistakenly describes dancing details that are not present at the end of the video. These errors often stem from a single step of misperception or flawed reasoning, yet they ultimately lead to incorrect final answers. In contrast, the direct answer provides an accurate and concise response for such perception-oriented tasks.

In Figure~\ref{fig:examples_videochat_fail_success}, we also show a success case of VideoChat-R1 on VideoMMMU. Unlike perception-oriented examples, this question involves a science problem based on an instructional video. In this context, the chain-of-thought reasoning process demonstrates a clear advantage: the model performs step-by-step deduction, correctly computes equations, and arrives at the final numerical result, which would be challenging via direct answering alone.

Next, we present qualitative results from \methodname{} across different benchmark types. In Figure~\ref{fig:examples_grounding}, we illustrate the model’s outputs on temporal grounding tasks. For these examples, the reasoning trace is typically straightforward—often limited to identifying when the action begins and ends. In many cases, the initial and reviewed answers are identical. Based on this observation, we apply early-exit directly on temporal grounding tasks without invoking further reasoning, which leads to reduced computation without sacrificing accuracy.

In Figure~\ref{fig:examples_qa}, we show results on perception-oriented QA benchmarks. For these relatively simple visual questions, \methodname{} consistently provides accurate responses in the initial answer, often accompanied by a high confidence score (\eg, over 99\%). These examples trigger early-exit behavior, allowing the model to maintain strong accuracy while improving inference efficiency.

In Figures~\ref{fig:examples_reasoning1}, we showcase examples from reasoning-intensive QA benchmarks. Compared to perception-oriented tasks, the reasoning traces here are significantly longer, with more detailed deduction steps. Notably, the model’s confidence in the initial answer is relatively low in such cases, allowing our confidence-based inference mechanism to trigger reasoning effectively.

\end{document}